\documentclass[sigconf]{acmart}
\AtBeginDocument{%
  \providecommand\BibTeX{{%
    \normalfont B\kern-0.5em{\scshape i\kern-0.25em b}\kern-0.8em\TeX}}}

\copyrightyear{2024}
\acmYear{2024}
\setcopyright{rightsretained}
\acmConference[CHI '24]{Proceedings of the CHI Conference on Human Factors in Computing Systems}{May 11--16, 2024}{Honolulu, HI, USA}
\acmBooktitle{Proceedings of the CHI Conference on Human Factors in Computing Systems (CHI '24), May 11--16, 2024, Honolulu, HI, USA}
\acmDOI{10.1145/3613904.3642570}
\acmISBN{979-8-4007-0330-0/24/05}

\usepackage{acmart-taps}
\usepackage{todonotes}
\usepackage{subcaption}

\DeclareMathOperator{\logit}{logit}

\newcommand{\ci}[3]{$\text{CI}_{#1\%} = [#2, #3]$}

\begin{document}

\title{Digital Comprehensibility Assessment of Simplified Texts among Persons with Intellectual Disabilities}

\author{Andreas Säuberli}
\email{andreas@cl.uzh.ch}
\orcid{0000-0001-9613-334X}
\affiliation{%
  \department{Department of Computational Linguistics}
  \institution{University of Zurich}
  \city{Zurich}
  \country{Switzerland}
}

\author{Franz Holzknecht}
\email{franz.holzknecht@hfh.ch}
\orcid{0000-0002-1218-2062}
\affiliation{%
  \department{Institute for Language and Communication}
  \institution{University of Teacher Education in Special Needs Zurich}
  \city{Zurich}
  \country{Switzerland}
}

\author{Patrick Haller}
\email{haller@cl.uzh.ch}
\orcid{0000-0002-8968-7587}
\affiliation{%
  \department{Department of Computational Linguistics}
  \institution{University of Zurich}
  \city{Zurich}
  \country{Switzerland}
}

\author{Silvana Deilen}
\email{sideilen@uni-mainz.de}
\orcid{0000-0002-2933-9257}
\affiliation{%
  \institution{Johannes Gutenberg University Mainz}
  \city{Germersheim}
  \country{Germany}
}

\author{Laura Schiffl}
\email{laura.schiffl@tum.de}
\orcid{0009-0003-4570-7472}
\affiliation{%
  \institution{Klinikum rechts der Isar, Technische Universität München}
  \city{München}
  \country{Germany}
}

\author{Silvia Hansen-Schirra}
\email{hansenss@uni-mainz.de}
\orcid{0000-0002-3277-7717}
\affiliation{%
  \institution{Johannes Gutenberg University Mainz}
  \city{Germersheim}
  \country{Germany}
}

\author{Sarah Ebling}
\email{ebling@cl.uzh.ch}
\orcid{0000-0001-6511-5085}
\affiliation{%
  \department{Department of Computational Linguistics}
  \institution{University of Zurich}
  \city{Zurich}
  \country{Switzerland}
}

\renewcommand{\shortauthors}{Säuberli et al.}

\begin{abstract}
Text simplification refers to the process of increasing the comprehensibility of texts. Automatic text simplification models are most commonly evaluated by experts or crowdworkers instead of the primary target groups of simplified texts, such as persons with intellectual disabilities. We conducted an evaluation study of text comprehensibility including participants with and without intellectual disabilities reading unsimplified, automatically and manually simplified German texts on a tablet computer. We explored four different approaches to measuring comprehensibility: multiple-choice comprehension questions, perceived difficulty ratings, response time, and reading speed. The results revealed significant variations in these measurements, depending on the reader group and whether the text had undergone automatic or manual simplification. For the target group of persons with intellectual disabilities, comprehension questions emerged as the most reliable measure, while analyzing reading speed provided valuable insights into participants' reading behavior.
\end{abstract}



\maketitle

\section{Introduction}
\label{sec:introduction}

Text simplification refers to the process of improving the comprehensibility of texts by reducing complexity at several linguistic levels, for instance, by using simpler vocabulary and syntactic structures, reorganizing text structure, and explaining difficult words and concepts. Primary target groups of simplified language\footnote{We use the term \emph{simplified language} as an umbrella term, including many (and often language-specific) varieties such as \emph{Easy Language} and \emph{Plain Language} \cite{Maass-2020}.} include persons with intellectual disabilities, persons with dementia, prelingually deaf persons, and non-native readers \cite{Maass-2020}. In recent years, both the demand for simplified texts and the amount of available data have been growing. Therefore, the development of quantitative human evaluation methods which include and represent the primary target groups becomes increasingly important. This is all the more true with increasing numbers of automatic text simplification (ATS) models being developed \cite{Thanyyan-Azmi-2021}. However, current research in ATS mostly resorts to evaluations based on opinions of experts (e.g., simplified language professionals) or crowdworkers who are not part of the primary target groups.

Meanwhile, the use of information and communication technology, and mobile touchscreen devices in particular, is becoming an integral part of the daily lives of persons in these target groups \cite{Ramsten-et-al-2018,Mendenez-et-al-2020}. This offers the potential of conducting human evaluations with target groups of simplified language in digital form. Apart from the efficiency gain in data collection and analysis compared to paper-and-pencil methods, digital assessment methods also allow participants to read texts in a more natural environment (possibly at home, on their own device), and make it possible to record detailed user interactions, enabling measurements such as reading speed or scrolling interactions as proxies for reading comprehension. However, there is currently little research on the most suitable methods for measuring text comprehensibility among the different target groups as well as on the effects of text simplification on these measurements. This issue also fundamentally concerns human-computer interaction, since persons with intellectual disabilities differ not only in their reading skills but also in their requirements for accessible user interfaces \cite{Braun-et-al-2020}.

The aim of the present study is to explore different ways of utilizing digital tools for measuring comprehensibility. We determine the \emph{comprehensibility} (sometimes also referred to as \emph{readability}) of a text by measuring its \emph{comprehension} on the part of members of a specific group of readers, while taking into account the fact that comprehensibility may differ between these groups. To discuss the suitability of these methods for evaluating ATS, we will also investigate the effect of the automatic simplification process on these measurements. More specifically, the study is guided by the following three research questions:

\begin{enumerate}
  \item Which methods for measuring comprehensibility can distinguish between simplified and non-simplified texts?
  \item What is the effect of manual and automatic text simplification on these measurements?
  \item How do these effects differ between persons with intellectual disabilities (as a primary target group of simplified language) and a control group of persons without intellectual disabilities?
\end{enumerate}

To answer these questions, we present results from an empirical study including participants with and without intellectual disabilities, using unsimplified, manually simplified (i.e., simplified by human experts), and automatically simplified German texts. To the best of our knowledge, this is the first study evaluating ATS for German with this target group.

\section{Related work}
\label{sec:related-work}

\subsection{Human evaluation of automatic text simplification}
\label{sec:related-work:human-evaluation}

While human evaluation is the preferred way of evaluating the quality of ATS output, there is no consensus on best practices \cite{Alva-Manchego-et-al-2020,Alva-Manchego-et-al-2021,Stajner-2021,Stodden-2021}. In recent ATS research where human evaluation was used, the most commonly applied methods were Likert scale ratings, usually for the categories \emph{simplicity}, \emph{fluency}/\emph{grammaticality}, and \emph{adequacy}/\emph{meaning preservation} \cite{Ryan-et-al-2023,Martin-et-al-2022,Stajner-Nisioi-2018,Mallinson-et-al-2020}. Less commonly, text comprehensibility or difficulty is evaluated using multiple-choice comprehension questions \cite{Alonzo-et-al-2021,Leroy-et-al-2013,Laban-et-al-2021} or free recall questions \cite{Leroy-et-al-2013}. Reading behavior, e.g., by measuring reading speed \cite{Alonzo-et-al-2021,Crossley-et-al-2014,Saggion-et-al-2015,Rello-et-al-2013b}, scrolling interactions \cite{Gooding-et-al-2021a}, or eye movements \cite{Rello-et-al-2013b}, is rarely considered.

In most cases, the participants of such comprehensibility studies are persons without disabilities or crowdworkers without specific inclusion criteria, who are not part of the primary target group of simplified language. This can be problematic, because what is considered difficult varies between reader groups \cite{Gooding-et-al-2021b,Stajner-2021}, and the requirements for text simplification should not be considered universal \cite{Gooding-2022}. Some exceptions of studies assessing ATS output among the target groups include experiments with deaf and hard-of-hearing adults \cite{Alonzo-et-al-2021}, persons with intellectual disabilities \cite{Huenerfauth-et-al-2009,Saggion-et-al-2015} or dyslexia \cite{Rello-et-al-2013b}, and language learners \cite{Crossley-et-al-2014}. Among these, \citet{Saggion-et-al-2015} is the most similar to our study, as they evaluated both manually and automatically simplified texts with persons with Down syndrome based on comprehension questions and reading time, in addition to an expert evaluation using Likert scale ratings. Their quantitative results did not show significant differences in comprehensibility between the different text versions, but they reported positive subjective perception of the simplified texts among target readers. Our study differs from this contribution in that it is fully digital, also making use of recorded user interactions, and we conduct the same comprehension assessment with persons with and without intellectual disabilities, which allows us to compare its effectiveness between the two groups.

\subsection{Comprehension of simplified language by persons with intellectual disabilities}
\label{sec:related-work:intellectual disabilities}

\citet{Fajardo-et-al-2014} conducted a study with 28 students with intellectual disability reading news articles in easy-to-read Spanish on paper, and correlated response accuracy in literal and inferential comprehension questions with linguistic measures such as word and sentence length. In a pilot study by \citet{Saletta-Winberg-2019}, 20 participants with intellectual or developmental disabilities read English texts that had undergone (among others) controlled manipulations reducing lexical and syntactic complexity. They measured errors while reading aloud and comprehension question response accuracy and found a significant effect on the former but not on the latter. They also found a high variability in reading comprehension among participants.

For German, several studies have investigated the effect of specific features of simplified language on comprehension by persons with intellectual disabilities. \citet{Schiffl-2020} conducted an experiment using eye-tracking with more than 80 participants, investigating the effects of word length and frequency. They found fundamental differences in eye movements while reading between persons with and without intellectual disabilities. \citet{Pappert-Bock-2019} studied compound segmentation (a feature in several varieties of simplified German) using a lexical decision task with participants with intellectual disability or functional illiteracy. \citet{Bock-Lange-2017} tested sentence and text comprehension skills of 28 persons with intellectual disabilities and showed that certain phenomena that are assumed to be too difficult for this target group (such as negation and personal pronouns) hardly caused any problems for the participants.

More generally, reading comprehension by target groups of simplified language has been studied by \citet{Jones-et-al-2006}, using several (adapted) standardized tests with participants with mild and borderline learning disabilities. This study revealed that the participant group was highly heterogeneous with respect to reading comprehension abilities.

\subsection{Digital assessment of reading comprehension}
\label{sec:related-work:digital-assessment}

As mentioned in Section \ref{sec:introduction}, digital assessment has the advantage of enabling measurements of reading behavior even without expensive equipment and expertise necessary for eye-tracking experiments. Some previous work has studied the connection between behavioral measurements such as reading speed or scrolling behavior and comprehension \cite{Dyson-Haselgrove-2000,Dyson-Haselgrove-2001,Wallot-et-al-2014,Stole-et-al-2020}, but there is only little research on exploiting these measurements for assessing comprehension or comprehensibility \cite{Gooding-et-al-2021a}. Our work contributes to this line of research by studying reading speed and response time as proxies for comprehension.

While there is a relatively large body of literature both in human-computer interaction and in language assessment dealing with differences in comprehension and behavior when reading on digital devices compared to paper \cite{AlSulaimi-AlShihi,Chen-Catrambone-2015,Kim-Kim-2013,Kong-et-al-2018,Saeuberli-et-al-2023}, almost no research has been conducted on how digital reading assessments need to be adapted for persons with intellectual disabilities. This is a significant research gap, given that these user groups have very different needs in terms of interface accessibility \cite{Braun-et-al-2020}. By comparing different assessment approaches between readers with and without intellectual disabilities, the present paper represents a first step towards addressing this research gap.

\section{Materials and methods}
\label{sec:methods}

\subsection{Texts and comprehension questions}
\label{sec:methods:texts}

The texts used in this study originate from a parallel corpus of original and simplified German documents. The documents were created at \emph{capito}, a provider of commercial text simplification services for German. Each document in the corpus was manually simplified by trained experts into one to three levels of simplification following the levels of the Common European Framework of Reference for Languages (A1, A2, and B2) \cite{cefr2020}. All manual simplifications used in this study are at level A2. This means that most of the information from the original text is retained (i.e., there is little to no summarization involved, as would be expected on a level of A1), but simpler syntactic structures and vocabulary are used, complex terminology is explained either inline or at the end of the text, and the layout is more readable, e.g., using bullet point lists and shorter line lengths. Level A2 is roughly comparable to Easy Language (in German: \emph{Leichte Sprache}), for which persons with intellectual disabilities are commonly listed as a primary target group \cite{Capito-LS,Bock-2014}.

We used a subset of this parallel corpus to train a neural ATS system (fine-tuned \emph{mBART} transformer model \cite{Liu-et-al-2020}) using the method described in \citet{Rios-et-al-2021}. From the remaining documents, we selected twelve texts according to several criteria: (1) The texts should be between 100 and 600 words in length, (2) they should cover a diverse range of topics but exclude topics known to be familiar to a wide audience, and (3) the texts should not require extensive additional context for comprehension. For each of the twelve documents, we generated an automatic simplification at the A2 level using the trained model and created four multiple-choice comprehension questions. The first question was always ``What is the text mainly about?'' (four answer options, one correct), the remaining three questions were about specific details present in the text (three answer options, one correct). We created these questions such that they can be answered based on the original and the manually simplified text, without looking at the automatic simplifications in order to avoid an unfair bias in favor of the system output. Since the ATS model sometimes erroneously omits information present in the original, the latter three questions have an additional fourth answer option ``Information does not appear in the text''. Care was taken that the questions are unambiguous, independent of each other (i.e., being able to answer a question was not contingent on getting the correct answer to a previous question), and unanswerable using world knowledge alone. Each question was double-checked for these criteria by two co-authors.

\subsection{Participants}
\label{sec:methods:participants}

We recruited two groups of participants from different populations, described in the following. All participants took part on a voluntary basis and were compensated monetarily.

\subsubsection{Target group}
\label{sec:methods:participants:target}

After approval by the institutional ethics review board, we recruited 18 participants from an educational program for persons with intellectual disabilities in Austria. Eight were female and ten were male, and they were aged between 18 and 32 (median: 23) at the time of recruitment. All participants had some form of cognitive impairment (most commonly: autism spectrum disorder, Down syndrome, or developmental delay), and a degree of disability of at least 50\% according to regulations concerning the assessment of the degree of disability in Austria\footnote{BGBl. II Nr. 261/2010, \url{https://www.ris.bka.gv.at/eli/bgbl/II/2010/261/20100818}}. Therefore, these participants represent a primary target group for simplified language. All participants were legally allowed to sign the consent forms themselves.

In a questionnaire, which all participants filled in before the first session, seven participants stated that they read texts in simplified language at least once per week, five at least once per month. A total of 17 stated that they used a touchscreen device on a daily basis, one person only weekly. Three participants did not list German as their native language, but all have completed compulsory education in German and are proficient at the CEFR level of A2 or higher.

\subsubsection{Control group}
\label{sec:methods:participants:control}

To compare the effects of text simplification on people outside the primary target groups, we additionally recruited 18 people without cognitive impairment---mostly current or former students---through university mailing lists. Twelve were female, six were male, and they were aged between 20 and 36 (median: 25). All were native German speakers.

Unlike in the target group, most participants in the control group were not used to reading simplified language (only 6 participants indicated reading simplified texts at least once per month). However, the information and consent forms which the participants received before the study were written in A2 simplified language to establish a basic level of familiarity.

\subsection{Procedure}
\label{sec:methods:procedure}

All experiments were conducted using the \emph{Okra} app (\cite{Saeuberli-et-al-2023}; version \verb|0.3.1-alpha|) on Apple iPads (9.7-inch). \emph{Okra} is an app for conducting reading experiments on mobile touchscreen devices, and it was specifically designed for and tested with users with intellectual disabilities. For instance, it reduces the complexity of the user interface and the amount of text on screen in order to decrease the cognitive load \cite{Saeuberli-et-al-2023}.

Each participant took part in three sessions on separate days. The target group sessions took place at the facilities of the educational program, the control group sessions in a university seminar room. Each session consisted of reading tasks, two sessions also included cognitive tasks. The app presented all instructions and guided participants through the entire session such that several participants could be tested simultaneously without interruptions. Each control group session included up to 12 participants, whereas for the target group, only up to 5 individuals participated per session. This was intended to provide better support in case of problems and to shorten waiting times, as reading speeds varied widely in the target group. One or two test administrators were present in the room and available for questions.

Before the main study, we conducted a usability test with 3 people from the same educational program to improve the usability and accessibility of the instructions and tasks implemented in the app. After finalizing the material and procedure, we piloted the entire experiment with 3 participants from the target group. Participants in the usability test and the pilot study were not recruited for the main study.

\subsubsection{Cognitive tasks}
\label{sec:methods:procedure:cognitive}

We included a total of four tasks testing several low-level cognitive skills related to reading. The purpose of these tasks was to provide a basic understanding of some of the differences between the two groups and the heterogeneity within each group. The tests we used were adaptations of tasks commonly used in psychological research (see references below). We adapted the tasks to the target group (by adjusting the difficulty and number of trials based on results from the usability test) and to the technical setup in the present study (by making the interface usable on a touchscreen).

\begin{itemize}
  \item \textbf{Digit span}: Memorizing and repeating an increasingly long sequence of digits in the same order \cite{Wechsler-2003}; two trials, each of which ended after two consecutive mistakes; measurement: longest correctly repeated sequence. This task tests short-term memory, sequencing ability, attention and automated learning \cite{Wechsler-2003}.
  \item \textbf{Lexical decision}: Deciding as quickly as possible whether the displayed strings of characters are words or pseudowords \cite{Meyer-Schvaneveldt-1971}; 37 stimuli; measurements: reaction time on correctly recognized words, ratio of correct responses. This task tests vocabulary knowledge and lexical access.
  \item \textbf{Reaction time}: Tapping randomly appearing balloons as quickly as possible; 15 stimuli; measurement: mean time between stimulus appearance and tap. Apart from motor aspects, reaction time also depends on cognitive factors such as visual processing speed and attention \citep{Amini-et-al-2019}.
  \item \textbf{Trail making}: Tapping randomly positioned numbers in ascending order as quickly as possible \cite{Reitan-Wolfson-1993}; 3 trials; measurement: mean time between taps. We only included part A of the trail making task, which primarily assesses visual attention and psychomotor speed \citep{Arnett-Labovitz-1995}.
\end{itemize}

Each task was preceded by a practice task, which participants could optionally repeat and whose results were excluded from the analysis.

\subsubsection{Reading tasks}
\label{sec:methods:procedure:reading}

\begin{figure*}
  \includegraphics[width=\textwidth]{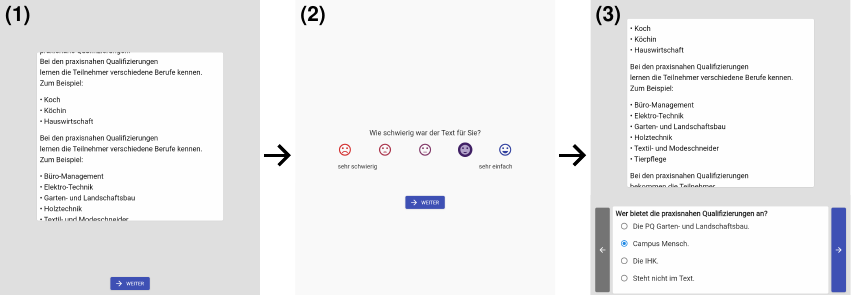}
  \Description[Screenshots of the reading task]{Three screenshots of the reading task. On the first screenshot, there is a German text in a scrollable white box in the center, surrounded by a gray background and a blue button at the bottom saying ``Continue'' in German. In the second screenshot, the question ``How difficult was the text for you?'' in German is shown in the center and five emoticons (frowning and red on the left, smiling and blue on the right); the leftmost emoticon is labeled ``very difficult'', the rightmost ``very easy''; beneath, there is a button saying ``Continue'' in German. In the third screenshot, the text is shown again in a white box, and a panel containing a comprehension question with four answer options next to radio buttons below.}
  \caption{Screenshots of the reading task in \emph{Okra}. \textbf{(1)} Initial reading screen, where only the text is visible. \textbf{(2)} Text difficulty rating screen. \textbf{(3)} Comprehension question screen; tapping the arrow buttons switches between questions.}
  \label{fig:screenshots}
\end{figure*}

Participants read four texts per session. The texts were presented in one of three versions (original, manually simplified, automatically simplified). No participant read the same text in more than one version. The design was counterbalanced so that all texts in all versions were read by the same number of participants in both groups. After reading the text, participants were asked to rate the text's difficulty on a 5-point scale (1 = \emph{very difficult}, 5 = \emph{very easy}), whereby the level descriptions were marked with textual labels, colors, and emoticons\footnote{We are aware that the interpretation of facial expressions, including emoticons, can differ between individuals. We used redundant coding with text, colors, and emoticons in the interface to avoid ambiguity, while also reducing the amount of text visible on screen.}. The text was then displayed again, along with the comprehension questions. Only one of four questions was shown at a time, and participants could switch back and forth between questions until they submitted their final answers. The screenshots in Figure \ref{fig:screenshots} show this procedure for one text. After finishing the text, participants were asked to take a break if necessary, and then continue with the next text.

Apart from the responses, we also recorded timestamped user interactions such as reading times and scrolling interactions. In the present paper, we will focus on the following measurements:

\begin{itemize}
  \item Responses to comprehension questions [with our assessment: correct/incorrect]
  \item Responses to text difficulty ratings [1--5]
  \item Time taken to answer each question, i.e., the total time during which the question was visible to the participant [seconds]
  \item Reading speed when initially reading the text [words per minute, WPM]
\end{itemize}

Since the ATS model sometimes does not transfer all information accurately and we designed the comprehension questions without looking at the ATS output, the correct answers in the automatic simplification could be different from the other versions. For example, the ATS model at times deleted a sentence from the original which included relevant information for answering a question, changing the correct answer for this question with respect to the automatically simplified text to ``Information does not appear in the text''. Therefore, we manually recoded the answer correctness for the automatically simplified texts. We removed instances where the correct answer in the automatic simplification was ``Information does not appear in the text'' in order not to give the ATS model an unfair advantage in the analysis. In total, we removed 9 out of 48 questions from the results of the automatically simplified texts.

\subsection{Statistical analysis}
\label{sec:methods:analysis}

When analyzing responses to comprehension questions or ratings, we took into account that some participants may be more proficient than others, and some questions may be more difficult to answer than others. To model these differences, we analyzed our data using the Rasch model, also called the one-parameter logistic model in item response theory (IRT). These models are widely used in language assessment and psychometrics, and formally comparable to generalized linear models with (fixed or random) effects for persons and items \cite[p.\ 143--145]{Fox-2010}\cite{DeBoeck-et-al-2011}. Whereas the classic Rasch model only considers persons and items in the analysis, the many-facets Rasch model allowed us to also take additional parameters (so-called \emph{facets}) into account in the modeling of the data \cite{Linacre-1989}. As we were interested in the effects of the text version (original, manually simplified, automatically simplified) on participants' performance, we  specified a many-facets Rasch model with three facets (persons, items, and text version). We used the estimated parameter values (the ``latent traits'') of the text version facet to compare the effect of manual and automatic text simplification.

We applied a dichotomous Rasch model for the comprehension questions \cite[p.\ 7--9]{Fox-2010} (equivalent to logistic regression) and a graded response model for the difficulty rating \cite{Samejima-1997}. For modeling response time and reading speed, we used log-linear regression models as in \cite{VanDerLinden-et-al-1999} and \cite[p.\ 228--231]{Fox-2010}, fitting person, item, and text version parameters in the same way as for the Rasch models. All models are defined in Table \ref{tab:models}.

\begin{table*}
  \centering
  \begin{tabular}{lll}
    \toprule
    Measurement & Range & Model definition \\
    \midrule
    Comprehension question accuracy & $y_{p,q,v} \in \{0,1\}$ & $P(y_{p,q,v}=1) = \logit^{-1}(\mu + \alpha_p - \beta_q - \delta_v)$ \\
    Perceived difficulty ratings & $y_{p,t,v} \in \{1,2,3,4,5\}$ & $\begin{aligned}P(y_{p,t,v}=k) = &\logit^{-1}(\mu + \alpha_p - \gamma_t - \delta_v - c_k) \\ & - \logit^{-1}(\mu + \alpha_p - \gamma_t - \delta_v - c_{k+1})\end{aligned}$ \\
    Response time & $y_{p,q,v} \in [0,\infty)$ & $\log(y_{p,q,v}) = \mu + \alpha_p + \beta_q + \delta_v + \epsilon$, $\epsilon \sim \mathcal{N}(0, \sigma)$ \\
    Reading speed & $y_{p,t,v} \in [0,\infty)$ & $\log(y_{p,t,v}) = \mu + \alpha_p + \gamma_t + \delta_v + \epsilon$, $\epsilon \sim \mathcal{N}(0, \sigma)$ \\
    \bottomrule
  \end{tabular}
  \caption{Model definitions for each measurement in the reading tasks. $\alpha_p$, $\beta_q$, $\gamma_t$, $\delta_v$ are parameters for a given person $p$, comprehension question $q$, text $t$, and text version $v$, $\mu$ and $\sigma$ are parameters for mean and standard deviation, and $c_k$ is a parameter for the threshold of each rating category $k$ (fixed to $-\infty$ for $k=1$).}
  \label{tab:models}
\end{table*}

We used Bayesian inference with a Markov chain Monte Carlo (MCMC) algorithm for fitting the models. This has several advantages compared to frequentist statistics: We get posterior distributions for parameter values, which provide more information than point estimates, it allows including prior knowledge, and Bayesian models are usually more accurate for complex IRT models and small sample sizes \cite[p.\ 2]{Fox-2010}\cite{Gao-Chen-2005}\cite{You-2022}. We defined wide normal distributions as priors for person, question/document, and text version parameters (cf.\ Table \ref{tab:priors}). For each measurement, we fit two separate models for the target and control groups, since we did not want to generalize across the different populations they are sampled from.

We used \emph{Stan} \cite{Carpenter-et-al-2017} with the \emph{PyStan} interface \cite{pystan} for sampling and \emph{ArviZ} \cite{Kumar-et-al-2019} for analysis. For MCMC, we used 4 chains with 2000 iterations, including 1000 warmup iterations. Model code and convergence diagnostics are published in the supplementary material.

\section{Results}
\label{sec:results}

Anonymized data and code for reproducing the analyses are available in the supplementary material. Five participants did not consent to publishing their anonymized raw data. Therefore, the data for these participants is not included in the supplementary material. The numbers and plots in the paper are based on the complete data.

\subsection{Cognitive tasks}
\label{sec:results:cognitive}

\begin{figure*}
  \centering
  \includegraphics[width=0.8\textwidth]{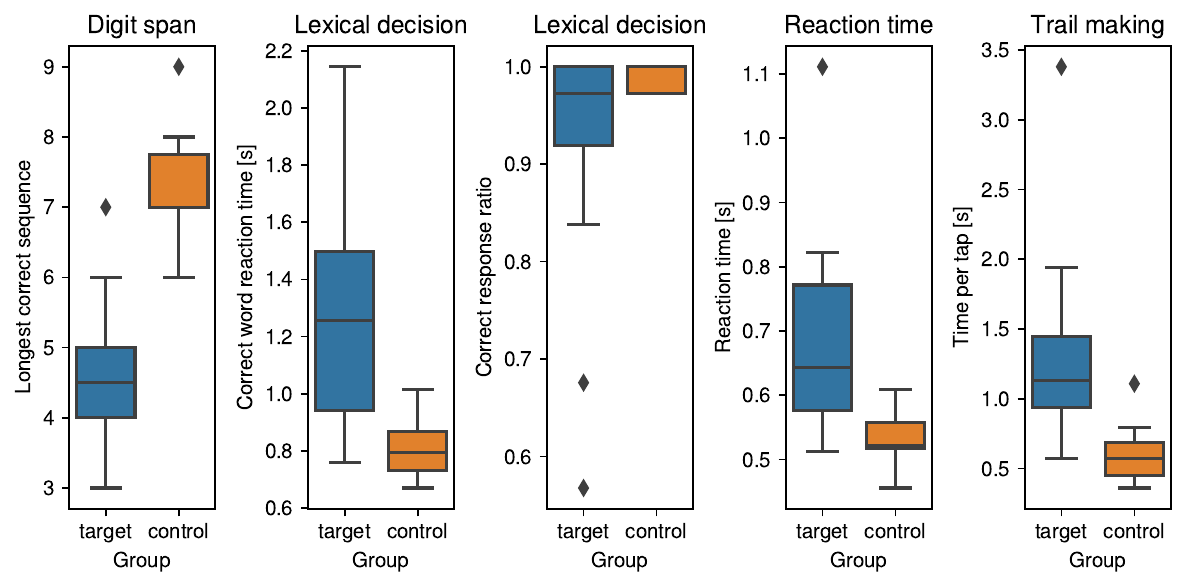}
  \Description[Boxplots of cognitive measurements]{Boxplots for each of the five measurements from the cognitive tasks, each comparing target and control group. First boxplot: longest correct sequence in the digit span task; interquartile ranges: target group between 4 and 5, control group between 7 and 7.75. Second boxplot: Correct word reaction time in seconds in the lexical decision task; interquartile ranges: target group between 0.94 and 1.50, control group between 0.73 and 0.87. Third boxplot: Correct response ratio in the lexical decision task; interquartile ranges: target group between 0.92 and 1.00, control group between 0.97 and 1.00. Fourth boxplot: Reaction time in seconds in the reaction time task; interquartile ranges: target group between 0.58 and 0.77, control group between 0.52 and 0.56. Fifth boxplot: Time per tap in seconds in the trail making task; interquartile ranges: target group between 0.94 and 1.45, control group between 0.45 and 0.68.}
  \caption{Boxplots of the measurements from the cognitive tasks, compared between target and control group. Each data point is the measured values for a single participant aggregated across all trials/stimuli (maximum for digit span, mean for all others), excluding practice trials.}
  \label{fig:cognitive}
\end{figure*}

Figure \ref{fig:cognitive} compares the measurements from the cognitive tasks between the target and control groups. The largest difference is in the digit span task for measuring working memory, with median scores of 4.5 for the target group and 7 for the control group. We also measured a longer reaction time in the lexical decision task, longer reaction times in general, and slower trail making in the target group. Moreover, variability in the target group is generally much higher than in the control group, which is likely to affect results in reading behavior and comprehension \cite{Johann-et-al-2020}.

\subsection{Reading tasks}
\label{sec:results:reading}

In total, 1680 responses to comprehension questions (excluding the 108 responses to unanswerable questions in the automatically simplified versions, see Section \ref{sec:methods:procedure:reading}) and 432 difficulty ratings are included in the analysis.

\begin{figure*}
  \hfill
  \begin{subfigure}[b]{0.48\textwidth}
    \includegraphics[width=\textwidth]{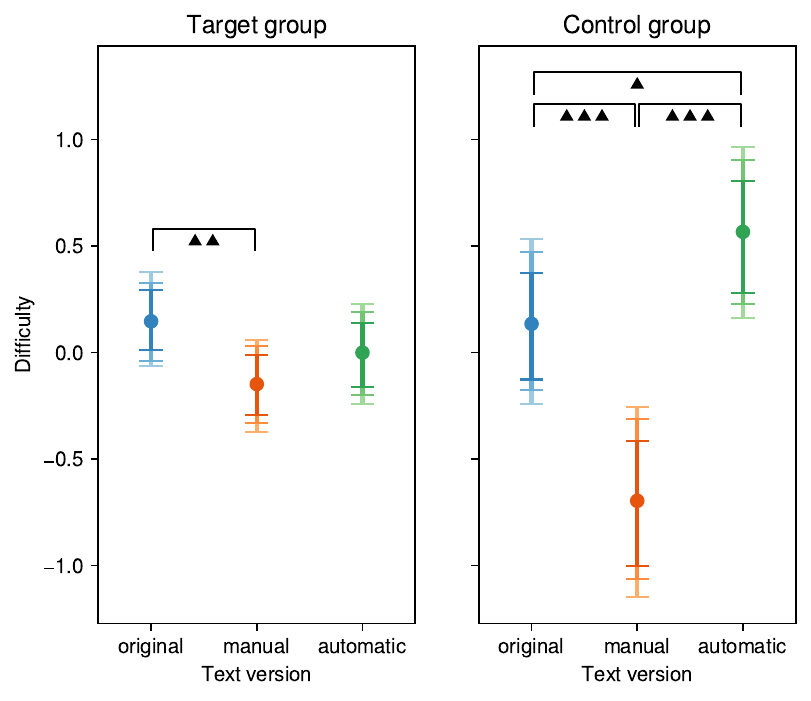}
    \Description[Point plots comparing comprehension question response accuracy between text versions]{Two point plots (one for the target group, one for the control group), each comparing the estimated difficulty between the three text versions, with error bars for credible intervals.}
    \caption{Comprehension question response accuracy.}
    \label{fig:reading:questions}
  \end{subfigure}
  \hfill
  \begin{subfigure}[b]{0.48\textwidth}
    \includegraphics[width=\textwidth]{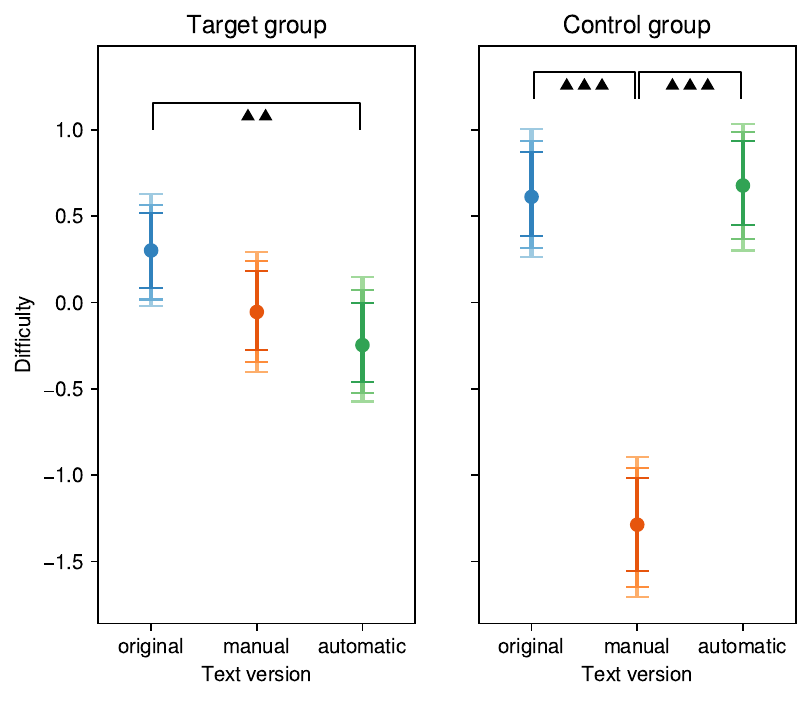}
    \Description[Point plots comparing perceived difficulty ratings between text versions]{Two point plots (one for the target group, one for the control group), each comparing the estimated difficulty between the three text versions, with error bars for credible intervals.}
    \caption{Perceived difficulty rating.}
    \label{fig:reading:ratings}
  \end{subfigure}
  \hfill
  \\[0.5cm]
  \hfill
  \begin{subfigure}[b]{0.48\textwidth}
    \includegraphics[width=\textwidth]{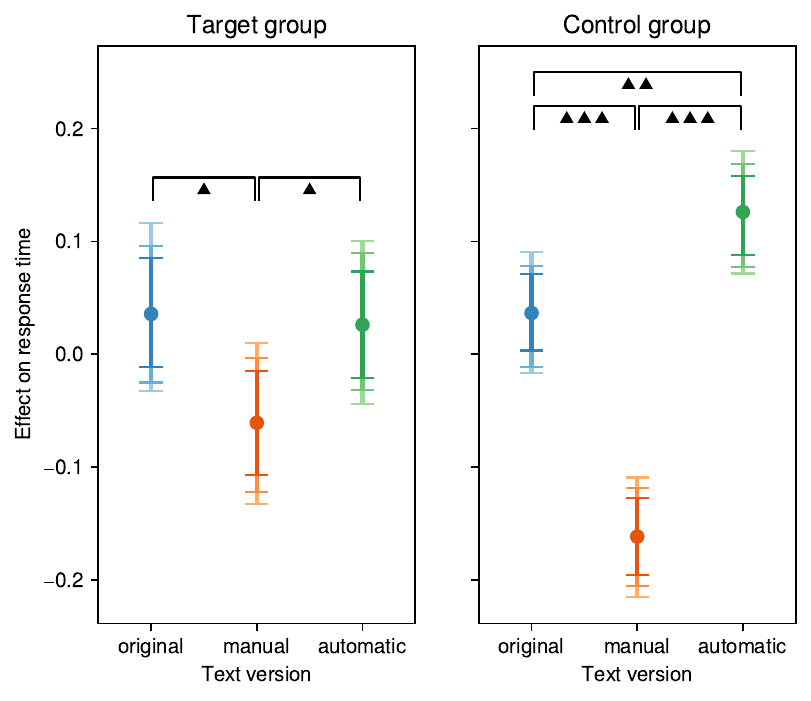}
    \Description[Point plots comparing comprehension question response time between text versions]{Two point plots (one for the target group, one for the control group), each comparing the estimated effect on response time between the three text versions, with error bars for credible intervals.}
    \caption{Comprehension question response time.}
    \label{fig:reading:response-times}
  \end{subfigure}
  \hfill
  \begin{subfigure}[b]{0.48\textwidth}
    \includegraphics[width=\textwidth]{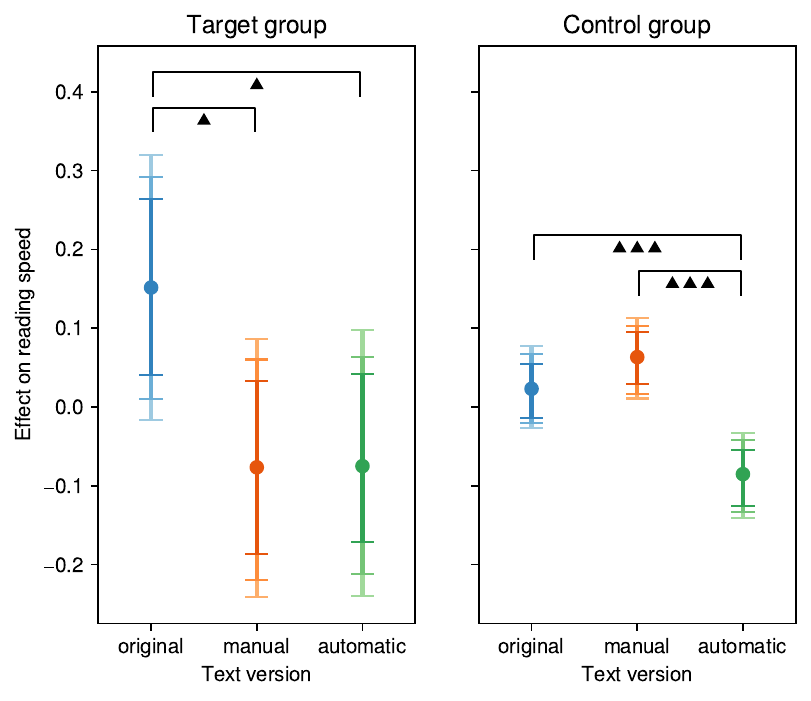}
    \Description[Point plots comparing reading speed between text versions]{Two point plots (one for the target group, one for the control group), each comparing the estimated effect on reading speed between the three text versions, with error bars for credible intervals.}
    \caption{Reading speed.}
    \label{fig:reading:speed}
  \end{subfigure}
  \hfill
  \caption{Posterior distributions of the text version parameters for the four measurements in the reading task. Points are medians, error bars are 80\%, 90\%, and 95\% credible intervals (CI). A bracket with $\blacktriangle$ indicates that the 80\% CI of the difference between the two parameters does not include zero (i.e., we are 80\% confident that there is a difference). Similarly with $\blacktriangle\blacktriangle$ for 90\% CI and $\blacktriangle\blacktriangle\blacktriangle$ for 95\% CI.}
  \label{fig:reading}
\end{figure*}

The estimated effects of the three text versions (original, manually simplified, automatically simplified) on the four measurements are visualized in Figure \ref{fig:reading}. Effects are centered around zero, and parameters for the two groups were estimated independently (as explained in Section \ref{sec:methods:analysis}), therefore the estimates cannot be compared across groups. We calculate the distribution of the difference between the three text version parameters at each MCMC sampling step and use highest density intervals (HDI) to quantify the credibility of the difference between the text version effects.

\subsubsection{Comprehension questions}
\label{sec:results:reading:questions}

Overall, the target group answered 47.5\% of the questions correctly, whereas the control group answered 92.8\% correctly. In the control group, 25 questions were answered correctly by all participants, and one participant answered all 48 questions correctly. In other words, in the control group, about half of the questions were uninformative because they were too easy and therefore unable to discriminate between more and less proficient readers and between more and less difficult text versions. This ceiling effect means that the parameter estimates of the Rasch model are less precise in the control group, as the wider credible intervals in Figure \ref{fig:reading:questions} show. Still, the estimated difficulty of the manual simplifications is measurably lower than both the original (\ci{95}{0.15}{1.61}) and the automatic simplifications (\ci{95}{0.49}{2.01}), meaning that participants had a significantly higher probability of answering questions correctly with the manually simplified version. The automatic simplifications appear to have been slightly more difficult than the originals (\ci{80}{0.01}{0.86}). In the target group, the effects are less pronounced, the original being the most difficult and the manual simplification the least difficult.

\subsubsection{Perceived difficulty ratings}
\label{sec:results:reading:ratings}

In Figure  \ref{fig:reading:ratings}, again, the differences between text versions are much smaller in the target group compared to the control group. The target group seems to rate the automatically simplified texts slightly easier than the originals (\ci{90}{0.08}{1.06}), whereas the control group rated the automatically simplified texts on par with the unsimplified ones. The control group had a strong tendency to rate the manually simplified texts as less difficult than the original (\ci{95}{1.22}{2.58}) and the automatically simplified texts (\ci{95}{1.26}{2.60}).

\subsubsection{Comprehension question response times}
\label{sec:results:reading:response-times}

In response time models, a larger effect means a longer response time, which is generally associated with a higher item difficulty in tests \cite{VanDerLinden-et-al-1999}. In the target group, from Figure \ref{fig:reading:response-times} we can observe that manual simplifications led to slightly faster response times (\ci{80}{0.02}{0.18}), while the automatic simplifications are on par with the originals. In the control group, the differences are even stronger, and the automatic simplifications appear to have been the most difficult. The effects on response time (Figure \ref{fig:reading:response-times}) look very similar to the effects on response accuracy (Figure \ref{fig:reading:questions}). This is in line with research on psychological research on test design \cite{VanDerLinden-et-al-1999}, but in our case, the observations from the two groups of participants do not agree on the relative difficulty of the automatically simplified texts.

\subsubsection{Reading speed}
\label{sec:results:reading:speed}

In terms of reading times, the behavior of the target group was much more variable and less predictable, as becomes obvious from Figure \ref{fig:reading:speed}. Some participants had implausible reading speeds of up to thousands of words per minute, meaning that many only skimmed or even skipped reading the text the first time it was displayed. We found that a small number of target group participants had a stronger tendency towards skimming or skipping, but most of them did not do so consistently, and reading speeds were not distributed bimodally, such that there was no obvious threshold to discriminate between reading and skipping. The slowest reading speeds (50 WPM and slower) were also observed in the target group. Mean reading speeds were 203 WPM in the target group and 168 WPM in the control group. For comparison, a standardized assessment of reading speed reported a mean of 179 WPM for native German speakers \cite{Trauzettel-Klosinski-et-al-2012}.

In the target group, the original texts tended to be read (or, more plausibly, skipped) more quickly than the two simplified versions (manual: \ci{80}{0.04}{0.42}, automatic: \ci{80}{0.05}{0.43}), while in the control group, the automatic simplifications were read more slowly than the other two (original: \ci{95}{0.02}{0.20}, manual: \ci{95}{0.06}{0.24}).

\section{Discussion}
\label{sec:discussion}

The primary goal of this study was to investigate four different measurement methods, comparing them with regards to two different text simplification methods and two different reader groups, with the ultimate aim of improving methods for human evaluation of ATS. We will discuss these aspects mainly based on the results in Figure \ref{fig:reading}. The purpose of the cognitive tasks was to characterize the participant groups to support the interpretation of results. Therefore, we will not discuss them in further detail here.

\subsection{Comparison of measurement methods and reader groups}
\label{sec:discussion:measurements-groups}

By design, there are several fundamental differences between the four measurement methods: Comprehension questions measure objective comprehension, while difficulty ratings measure subjective perception. Measurements such as response time and reading speed can only serve as proxies for comprehension and require specific assumptions about the behavior of participants. For any of these measurements to be considered suitable for evaluating text simplification, they need to be able to capture a difference between less comprehensible and more comprehensible texts. Since the manually simplified texts were professionally edited by trained experts and according to guidelines developed and checked with target readers, it is safe to assume that there should be some measurable difference in comprehensibility between original and manually simplified texts. From this perspective, our results suggest that the measurement of comprehension question response accuracy was most successful, and perceived difficulty ratings were least successful with the target group. For the control group, all measurements except reading speed were successful in differentiating between original and manually simplified texts.

There are two possible factors which may explain why ratings were less reliable than comprehension questions for the target group: First, the target group was quite heterogeneous (as evidenced by the cognitive tasks), which led to larger differences in subjective judgments of texts, especially because we did not give more specific instructions to calibrate ratings in order to reduce cognitive load. Second, when readers lose motivation and stop reading, which happened in the target group, rating responses may be more random, whereas responses to comprehension questions will reliably show a random-guessing accuracy. Both of these may be arguments against using perceived difficulty ratings with the target group.

Familiarity is a confounding factor, because participants in the target group were mostly very familiar with the specific variety of simplified language in the study, which may have biased their perception of the texts. However, if this bias was strong, we would expect a larger effect in the perceived difficulty ratings compared to the control group participants, who were mostly unfamiliar with simplified language.

While previous work heavily relied on ratings for evaluating the comprehensibility of simplified texts (see Section \ref{sec:related-work:human-evaluation}), our results show that this is not always sufficient, especially for readers with intellectual disabilities. The results also revealed significant differences in comprehensibility and perception between persons with and without intellectual disabilities, highlighting the importance of including the primary target groups in studies on text simplification and bridging the gap to insights from psycholinguistic research (see Section \ref{sec:related-work:intellectual disabilities}).

Although reading speed was mostly unsuccessful in discriminating between text versions, it revealed important behavioral patterns in the target group (skimming/skipping texts), which supports the interpretation of other results. Previous work has suggested that reading time is to be considered separately from comprehension \cite{Wallot-et-al-2014}. Our observations support this view, but our interpretation is limited by our study design: Since the text was shown again after the initial reading, participants were free to decide not to read the entire text the first time around.

Overall, the fact that reading behavior can be measured through a mobile application is a major advantage of using digital evaluation tools such as the one described in this paper compared to paper-and-pencil assessment. Our work represents a first step towards exploiting this advantage to make comprehensibility assessment more inclusive (see Section \ref{sec:related-work:digital-assessment}).

\subsection{Effect of automatic simplification}

We have seen that manual simplification resulted in noticeable differences for most measures. By comparing the difficulty estimates in the automatic simplification to the original and manual simplification, we can evaluate the ATS output in terms of comprehensibility.

Based on the target group measurements, automatic simplification only had a modest effect. The largest improvement compared to the original texts was observed in the perceived difficulty ratings, which were generally less reliable with this group, as discussed in Section \ref{sec:discussion:measurements-groups}. However, in terms of reading speed, ATS had the same effect as manual simplification, which suggests that ATS was somewhat successful in keeping up motivation to continue reading for the target group. A possible explanation for this is that at the surface level, the texts looked more like the simplified texts the participants were familiar with.

In the control group, all measurements agreed that ATS outputs were equally difficult or more difficult than the original text. Apart from lack of quality in the automatic simplifications, several factors may have contributed to these results: First, as described in Section \ref{sec:methods:texts}, the comprehension questions were written and optimized for the original and manually simplified texts. Although we removed responses to questions which were not answerable based on the ATS output, the wording of the questions may still have made the questions more difficult in the automatic simplification. Second, the control group may be more perceptive or sensitive to grammatical and semantic errors in the text than the target group. Evidence for this are the control group's higher difficulty ratings and lower reading speed for the automatically simplified texts.

Particularly the second factor requires further experimental research, as it could have a major influence on human evaluation of text simplification: If the linguistic fluency of ATS output only has a weak influence on comprehension in primary target groups of simplified texts, this should be accounted for when evaluating ATS systems. In addition, different types of linguistic errors may have a different effect on reading behavior depending on the specific type of cognitive impairment of the reader \cite{Micai-et-al-2019}.

In the present work, we focused on the evaluation methodology as opposed to pinpointing specific problems in the ATS output. However, our findings on the comprehensibility of automatically simplified texts are in line with current research showing that ATS systems are still quite limited in the effective gain of simplicity they can achieve \cite{Ryan-et-al-2023}. Recent experiments with large instruction-tuned language models have already suggested significant improvements in this regard, and these models would likely outperform our models \cite{Maddela-et-al-2023,Kew-et-al-2023}. It is all the more important that these improvements are evaluated with primary target reader groups in the future.

\section{Conclusion}
\label{sec:conclusion}

We conducted a study exploring different ways of measuring text comprehensibility using a mobile application and investigating the effect of manual and automatic text simplification on comprehension, including participants with and without intellectual disabilities. The results revealed several types of differences which must be taken into account when designing human evaluation studies:

\begin{itemize}
  \item Differences between \textbf{measurement methods}: Comprehension questions, difficulty ratings, and behavioral measurements can lead to different conclusions and complement each other when combined.
  \item Differences between \textbf{manual and automatic simplification}: Issues in the ATS output may significantly impair objective comprehension without affecting subjective perception (especially in the target group), whereas manually simplified texts lead to more predictable results across measurement methods.
  \item Differences between \textbf{reader groups}: Results from persons with intellectual disabilities can be different from (or even in contradiction to) those of persons without disabilities, particularly in terms of reading behavior and subjective perception of difficulty.
\end{itemize}

We consider measuring interactions of users  reading on touchscreen devices to be a promising approach, especially for assessing comprehensibility with diverse target groups, as traditional tests based on comprehension questions can be cognitively demanding. Another advantage is that this approach allows assessing reading behavior in a more natural environment. However, further research is still required on other aspects of human-computer interaction, e.g., regarding the exact relationship between user interactions and text comprehension, the ways in which interactions differ between persons with and without intellectual disabilities, and how they can be used to design more reliable and accessible comprehensibility assessments with diverse user groups.

Overall, we show that applying digital assessment methods for comprehensibility evaluation to persons with intellectual disabilities is viable, and that combining subjective and objective comprehensibility assessment with behavioral measurements provides valuable insights into the impact of text simplification.

\begin{acks}
We are greatly indebted to CFS GmbH (capito), in particular to Ursula Semlitsch, for recruiting and running the sessions with the target group participants, and helping to translate the material into simplified language. We would also like to thank Silke Gutermuth for helpful feedback on the experiment design and the comprehension questions. We thank the anonymous reviewers for their constructive comments. This work was funded by the Swiss Innovation Agency (Innosuisse) Flagship IICT (PFFS-21-47).
\end{acks}

\bibliographystyle{ACM-Reference-Format}
\bibliography{references}


\begin{thebibliography}{63}


\ifx \showCODEN    \undefined \def \showCODEN     #1{\unskip}     \fi
\ifx \showDOI      \undefined \def \showDOI       #1{#1}\fi
\ifx \showISBNx    \undefined \def \showISBNx     #1{\unskip}     \fi
\ifx \showISBNxiii \undefined \def \showISBNxiii  #1{\unskip}     \fi
\ifx \showISSN     \undefined \def \showISSN      #1{\unskip}     \fi
\ifx \showLCCN     \undefined \def \showLCCN      #1{\unskip}     \fi
\ifx \shownote     \undefined \def \shownote      #1{#1}          \fi
\ifx \showarticletitle \undefined \def \showarticletitle #1{#1}   \fi
\ifx \showURL      \undefined \def \showURL       {\relax}        \fi
\providecommand\bibfield[2]{#2}
\providecommand\bibinfo[2]{#2}
\providecommand\natexlab[1]{#1}
\providecommand\showeprint[2][]{arXiv:#2}

\bibitem[Al-Sulaimi and Al-Shihi(2017)]%
        {AlSulaimi-AlShihi}
\bibfield{author}{\bibinfo{person}{Aisha Al-Sulaimi} {and}
  \bibinfo{person}{Hafedh Al-Shihi}.} \bibinfo{year}{2017}\natexlab{}.
\newblock \showarticletitle{The effects of reading mode (digital vs printed
  text) on reading comprehension: A literature review of the key assessment
  factors}. In \bibinfo{booktitle}{\emph{2017 6th International Conference on
  Information and Communication Technology and Accessibility (ICTA)}}.
  \bibinfo{pages}{1--6}.
\newblock
\urldef\tempurl%
\url{https://doi.org/10.1109/ICTA.2017.8336053}
\showDOI{\tempurl}


\bibitem[Al-Thanyyan and Azmi(2021)]%
        {Thanyyan-Azmi-2021}
\bibfield{author}{\bibinfo{person}{Suha~S. Al-Thanyyan} {and}
  \bibinfo{person}{Aqil~M. Azmi}.} \bibinfo{year}{2021}\natexlab{}.
\newblock \showarticletitle{Automated Text Simplification: A Survey}.
\newblock \bibinfo{journal}{\emph{ACM Comput. Surv.}} \bibinfo{volume}{54},
  \bibinfo{number}{2}, Article \bibinfo{articleno}{43} (\bibinfo{date}{mar}
  \bibinfo{year}{2021}), \bibinfo{numpages}{36}~pages.
\newblock
\showISSN{0360-0300}
\urldef\tempurl%
\url{https://doi.org/10.1145/3442695}
\showDOI{\tempurl}


\bibitem[Alonzo et~al\mbox{.}(2021)]%
        {Alonzo-et-al-2021}
\bibfield{author}{\bibinfo{person}{Oliver Alonzo}, \bibinfo{person}{Jessica
  Trussell}, \bibinfo{person}{Becca Dingman}, {and} \bibinfo{person}{Matt
  Huenerfauth}.} \bibinfo{year}{2021}\natexlab{}.
\newblock \showarticletitle{Comparison of Methods for Evaluating Complexity of
  Simplified Texts among Deaf and Hard-of-Hearing Adults at Different Literacy
  Levels}. In \bibinfo{booktitle}{\emph{Proceedings of the 2021 CHI Conference
  on Human Factors in Computing Systems}}. \bibinfo{pages}{1--12}.
\newblock


\bibitem[Alva-Manchego et~al\mbox{.}(2020)]%
        {Alva-Manchego-et-al-2020}
\bibfield{author}{\bibinfo{person}{Fernando Alva-Manchego},
  \bibinfo{person}{Carolina Scarton}, {and} \bibinfo{person}{Lucia Specia}.}
  \bibinfo{year}{2020}\natexlab{}.
\newblock \showarticletitle{Data-Driven Sentence Simplification: Survey and
  Benchmark}.
\newblock \bibinfo{journal}{\emph{Computational Linguistics}}
  \bibinfo{volume}{46}, \bibinfo{number}{1} (\bibinfo{year}{2020}),
  \bibinfo{pages}{135--187}.
\newblock
\urldef\tempurl%
\url{https://doi.org/10.1162/coli_a_00370}
\showDOI{\tempurl}


\bibitem[Alva-Manchego et~al\mbox{.}(2021)]%
        {Alva-Manchego-et-al-2021}
\bibfield{author}{\bibinfo{person}{Fernando Alva-Manchego},
  \bibinfo{person}{Carolina Scarton}, {and} \bibinfo{person}{Lucia Specia}.}
  \bibinfo{year}{2021}\natexlab{}.
\newblock \showarticletitle{The (Un)Suitability of Automatic Evaluation Metrics
  for Text Simplification}.
\newblock \bibinfo{journal}{\emph{Computational Linguistics}}
  \bibinfo{volume}{47}, \bibinfo{number}{4} (\bibinfo{date}{Dec.}
  \bibinfo{year}{2021}), \bibinfo{pages}{861--889}.
\newblock
\urldef\tempurl%
\url{https://doi.org/10.1162/coli_a_00418}
\showDOI{\tempurl}


\bibitem[Amini~Vishteh et~al\mbox{.}(2019)]%
        {Amini-et-al-2019}
\bibfield{author}{\bibinfo{person}{Rasoul Amini~Vishteh}, \bibinfo{person}{Ali
  Mirzajani}, \bibinfo{person}{Ebrahim Jafarzadehpour}, {and}
  \bibinfo{person}{Samireh Darvishpour}.} \bibinfo{year}{2019}\natexlab{}.
\newblock \showarticletitle{Evaluation of simple visual reaction time of
  different colored light stimuli in visually normal students}.
\newblock \bibinfo{journal}{\emph{Clinical Optometry}} (\bibinfo{year}{2019}),
  \bibinfo{pages}{167--171}.
\newblock


\bibitem[Arnett and Labovitz(1995)]%
        {Arnett-Labovitz-1995}
\bibfield{author}{\bibinfo{person}{James~A Arnett} {and}
  \bibinfo{person}{Seth~S Labovitz}.} \bibinfo{year}{1995}\natexlab{}.
\newblock \showarticletitle{Effect of physical layout in performance of the
  Trail Making Test.}
\newblock \bibinfo{journal}{\emph{Psychological Assessment}}
  \bibinfo{volume}{7}, \bibinfo{number}{2} (\bibinfo{year}{1995}),
  \bibinfo{pages}{220}.
\newblock


\bibitem[Bock(2014)]%
        {Bock-2014}
\bibfield{author}{\bibinfo{person}{Bettina~M. Bock}.}
  \bibinfo{year}{2014}\natexlab{}.
\newblock \bibinfo{booktitle}{\emph{“Leichte Sprache”: Abgrenzung,
  Beschreibung und Problemstellungen aus Sicht der Linguistik}}.
\newblock \bibinfo{publisher}{Frank \& Timme}, \bibinfo{address}{Berlin},
  \bibinfo{pages}{17--51}.
\newblock


\bibitem[Bock and Lange(2017)]%
        {Bock-Lange-2017}
\bibfield{author}{\bibinfo{person}{Bettina~M. Bock} {and}
  \bibinfo{person}{Daisy Lange}.} \bibinfo{year}{2017}\natexlab{}.
\newblock \bibinfo{booktitle}{\emph{Empirische Untersuchungen zu Satz- und
  Textverstehen bei Menschen mit geistiger Behinderung und funktionalen
  Analphabeten}}.
\newblock \bibinfo{publisher}{Frank \& Timme}, \bibinfo{address}{Berlin},
  \bibinfo{pages}{253--274}.
\newblock


\bibitem[Braun et~al\mbox{.}(2020)]%
        {Braun-et-al-2020}
\bibfield{author}{\bibinfo{person}{Melinda Braun}, \bibinfo{person}{Matthias
  Wölfel}, \bibinfo{person}{Gregor Renner}, {and} \bibinfo{person}{Christian
  Menschik}.} \bibinfo{year}{2020}\natexlab{}.
\newblock \showarticletitle{Accessibility of Different Natural User Interfaces
  for People with Intellectual Disabilities}. In \bibinfo{booktitle}{\emph{2020
  International Conference on Cyberworlds (CW)}}. \bibinfo{pages}{211--218}.
\newblock
\urldef\tempurl%
\url{https://doi.org/10.1109/CW49994.2020.00041}
\showDOI{\tempurl}


\bibitem[Capito(2023)]%
        {Capito-LS}
\bibfield{author}{\bibinfo{person}{Capito}.} \bibinfo{year}{2023}\natexlab{}.
\newblock \bibinfo{title}{{E}asy {L}anguage: What is it and why is it
  important?}
\newblock
\newblock
\urldef\tempurl%
\url{https://capito.eu/en/easy-language/}
\showURL{%
\tempurl}
\newblock
\shownote{Accessed: 2023-12-07}.


\bibitem[Carpenter et~al\mbox{.}(2017)]%
        {Carpenter-et-al-2017}
\bibfield{author}{\bibinfo{person}{Bob Carpenter}, \bibinfo{person}{Andrew
  Gelman}, \bibinfo{person}{Matthew~D. Hoffman}, \bibinfo{person}{Daniel Lee},
  \bibinfo{person}{Ben Goodrich}, \bibinfo{person}{Michael Betancourt},
  \bibinfo{person}{Marcus Brubaker}, \bibinfo{person}{Jiqiang Guo},
  \bibinfo{person}{Peter Li}, {and} \bibinfo{person}{Allen Riddell}.}
  \bibinfo{year}{2017}\natexlab{}.
\newblock \showarticletitle{Stan: A Probabilistic Programming Language}.
\newblock \bibinfo{journal}{\emph{Journal of Statistical Software}}
  \bibinfo{volume}{76}, \bibinfo{number}{1} (\bibinfo{year}{2017}),
  \bibinfo{pages}{1--32}.
\newblock
\urldef\tempurl%
\url{https://doi.org/10.18637/jss.v076.i01}
\showDOI{\tempurl}


\bibitem[Chen and Catrambone(2015)]%
        {Chen-Catrambone-2015}
\bibfield{author}{\bibinfo{person}{Dar-Wei Chen} {and} \bibinfo{person}{Richard
  Catrambone}.} \bibinfo{year}{2015}\natexlab{}.
\newblock \showarticletitle{Paper vs. Screen: Effects on Reading Comprehension,
  Metacognition, and Reader Behavior}.
\newblock \bibinfo{journal}{\emph{Proceedings of the Human Factors and
  Ergonomics Society Annual Meeting}} \bibinfo{volume}{59}, \bibinfo{number}{1}
  (\bibinfo{year}{2015}), \bibinfo{pages}{332--336}.
\newblock
\urldef\tempurl%
\url{https://doi.org/10.1177/1541931215591069}
\showDOI{\tempurl}
\showeprint{https://doi.org/10.1177/1541931215591069}


\bibitem[{Council of Europe}(2020)]%
        {cefr2020}
\bibfield{author}{\bibinfo{person}{{Council of Europe}}.}
  \bibinfo{year}{2020}\natexlab{}.
\newblock \bibinfo{booktitle}{\emph{{Common {European} framework of reference
  for languages: {Learning}, teaching, assessment. {Companion} volume}}}.
\newblock \bibinfo{publisher}{Council of Europe Publishing},
  \bibinfo{address}{Strasbourg}.
\newblock


\bibitem[Crossley et~al\mbox{.}(2014)]%
        {Crossley-et-al-2014}
\bibfield{author}{\bibinfo{person}{Scott~A Crossley}, \bibinfo{person}{Hae~Sung
  Yang}, {and} \bibinfo{person}{Danielle~S McNamara}.}
  \bibinfo{year}{2014}\natexlab{}.
\newblock \showarticletitle{What's so Simple about Simplified Texts? A
  Computational and Psycholinguistic Investigation of Text Comprehension and
  Text Processing.}
\newblock \bibinfo{journal}{\emph{Reading in a Foreign Language}}
  \bibinfo{volume}{26}, \bibinfo{number}{1} (\bibinfo{year}{2014}),
  \bibinfo{pages}{92--113}.
\newblock


\bibitem[De~Boeck et~al\mbox{.}(2011)]%
        {DeBoeck-et-al-2011}
\bibfield{author}{\bibinfo{person}{Paul De~Boeck}, \bibinfo{person}{Marjan
  Bakker}, \bibinfo{person}{Robert Zwitser}, \bibinfo{person}{Michel Nivard},
  \bibinfo{person}{Abe Hofman}, \bibinfo{person}{Francis Tuerlinckx}, {and}
  \bibinfo{person}{Ivailo Partchev}.} \bibinfo{year}{2011}\natexlab{}.
\newblock \showarticletitle{The Estimation of Item Response Models with the
  lmer Function from the lme4 Package in R}.
\newblock \bibinfo{journal}{\emph{Journal of Statistical Software}}
  \bibinfo{volume}{39}, \bibinfo{number}{12} (\bibinfo{year}{2011}),
  \bibinfo{pages}{1--28}.
\newblock
\urldef\tempurl%
\url{https://doi.org/10.18637/jss.v039.i12}
\showDOI{\tempurl}


\bibitem[Dyson and Haselgrove(2000)]%
        {Dyson-Haselgrove-2000}
\bibfield{author}{\bibinfo{person}{Mary Dyson} {and} \bibinfo{person}{Mark
  Haselgrove}.} \bibinfo{year}{2000}\natexlab{}.
\newblock \showarticletitle{The effects of reading speed and reading patterns
  on the understanding of text read from screen}.
\newblock \bibinfo{journal}{\emph{Journal of Research in Reading}}
  \bibinfo{volume}{23}, \bibinfo{number}{2} (\bibinfo{year}{2000}),
  \bibinfo{pages}{210--223}.
\newblock
\urldef\tempurl%
\url{https://doi.org/10.1111/1467-9817.00115}
\showDOI{\tempurl}
\showeprint{https://onlinelibrary.wiley.com/doi/pdf/10.1111/1467-9817.00115}


\bibitem[DYSON and HASELGROVE(2001)]%
        {Dyson-Haselgrove-2001}
\bibfield{author}{\bibinfo{person}{MARY~C. DYSON} {and} \bibinfo{person}{MARK
  HASELGROVE}.} \bibinfo{year}{2001}\natexlab{}.
\newblock \showarticletitle{The influence of reading speed and line length on
  the effectiveness of reading from screen}.
\newblock \bibinfo{journal}{\emph{International Journal of Human-Computer
  Studies}} \bibinfo{volume}{54}, \bibinfo{number}{4} (\bibinfo{year}{2001}),
  \bibinfo{pages}{585--612}.
\newblock
\showISSN{1071-5819}
\urldef\tempurl%
\url{https://doi.org/10.1006/ijhc.2001.0458}
\showDOI{\tempurl}


\bibitem[Fajardo et~al\mbox{.}(2014)]%
        {Fajardo-et-al-2014}
\bibfield{author}{\bibinfo{person}{Inmaculada Fajardo},
  \bibinfo{person}{Vicenta {\'A}vila}, \bibinfo{person}{Antonio Ferrer},
  \bibinfo{person}{Gema Tavares}, \bibinfo{person}{Marcos G{\'o}mez}, {and}
  \bibinfo{person}{Ana Hern{\'a}ndez}.} \bibinfo{year}{2014}\natexlab{}.
\newblock \showarticletitle{Easy-to-read texts for students with intellectual
  disability: linguistic factors affecting comprehension}.
\newblock \bibinfo{journal}{\emph{Journal of applied research in intellectual
  disabilities}} \bibinfo{volume}{27}, \bibinfo{number}{3}
  (\bibinfo{year}{2014}), \bibinfo{pages}{212--225}.
\newblock


\bibitem[Fox(2010)]%
        {Fox-2010}
\bibfield{author}{\bibinfo{person}{Jean-Paul Fox}.}
  \bibinfo{year}{2010}\natexlab{}.
\newblock \bibinfo{booktitle}{\emph{Bayesian Item Response Modeling: Theory and
  Applications}}.
\newblock \bibinfo{publisher}{Springer New York}, \bibinfo{address}{New York,
  NY}.
\newblock
\urldef\tempurl%
\url{https://doi.org/10.1007/978-1-4419-0742-4}
\showDOI{\tempurl}


\bibitem[Gao and Chen(2005)]%
        {Gao-Chen-2005}
\bibfield{author}{\bibinfo{person}{Furong Gao} {and} \bibinfo{person}{Lisue
  Chen}.} \bibinfo{year}{2005}\natexlab{}.
\newblock \showarticletitle{Bayesian or Non-Bayesian: A Comparison Study of
  Item Parameter Estimation in the Three-Parameter Logistic Model}.
\newblock \bibinfo{journal}{\emph{Applied Measurement in Education}}
  \bibinfo{volume}{18}, \bibinfo{number}{4} (\bibinfo{date}{oct}
  \bibinfo{year}{2005}), \bibinfo{pages}{351--380}.
\newblock
\urldef\tempurl%
\url{https://doi.org/10.1207/s15324818ame1804_2}
\showDOI{\tempurl}


\bibitem[Gooding(2022)]%
        {Gooding-2022}
\bibfield{author}{\bibinfo{person}{Sian Gooding}.}
  \bibinfo{year}{2022}\natexlab{}.
\newblock \showarticletitle{On the Ethical Considerations of Text
  Simplification}. In \bibinfo{booktitle}{\emph{Ninth Workshop on Speech and
  Language Processing for Assistive Technologies (SLPAT-2022)}}.
  \bibinfo{publisher}{Association for Computational Linguistics},
  \bibinfo{address}{Dublin, Ireland}, \bibinfo{pages}{50--57}.
\newblock
\urldef\tempurl%
\url{https://doi.org/10.18653/v1/2022.slpat-1.7}
\showDOI{\tempurl}


\bibitem[Gooding et~al\mbox{.}(2021a)]%
        {Gooding-et-al-2021a}
\bibfield{author}{\bibinfo{person}{Sian Gooding}, \bibinfo{person}{Yevgeni
  Berzak}, \bibinfo{person}{Tony Mak}, {and} \bibinfo{person}{Matt Sharifi}.}
  \bibinfo{year}{2021}\natexlab{a}.
\newblock \showarticletitle{Predicting Text Readability from Scrolling
  Interactions}. In \bibinfo{booktitle}{\emph{Proceedings of the 25th
  Conference on Computational Natural Language Learning}}.
  \bibinfo{publisher}{Association for Computational Linguistics},
  \bibinfo{address}{Online}, \bibinfo{pages}{380--390}.
\newblock
\urldef\tempurl%
\url{https://doi.org/10.18653/v1/2021.conll-1.30}
\showDOI{\tempurl}


\bibitem[Gooding et~al\mbox{.}(2021b)]%
        {Gooding-et-al-2021b}
\bibfield{author}{\bibinfo{person}{Sian Gooding}, \bibinfo{person}{Ekaterina
  Kochmar}, \bibinfo{person}{Seid~Muhie Yimam}, {and} \bibinfo{person}{Chris
  Biemann}.} \bibinfo{year}{2021}\natexlab{b}.
\newblock \showarticletitle{Word Complexity is in the Eye of the Beholder}. In
  \bibinfo{booktitle}{\emph{Proceedings of the 2021 Conference of the North
  American Chapter of the Association for Computational Linguistics: Human
  Language Technologies}}. \bibinfo{publisher}{Association for Computational
  Linguistics}, \bibinfo{address}{Online}, \bibinfo{pages}{4439--4449}.
\newblock
\urldef\tempurl%
\url{https://doi.org/10.18653/v1/2021.naacl-main.351}
\showDOI{\tempurl}


\bibitem[Huenerfauth et~al\mbox{.}(2009)]%
        {Huenerfauth-et-al-2009}
\bibfield{author}{\bibinfo{person}{Matt Huenerfauth}, \bibinfo{person}{Lijun
  Feng}, {and} \bibinfo{person}{No{\'e}mie Elhadad}.}
  \bibinfo{year}{2009}\natexlab{}.
\newblock \showarticletitle{Comparing evaluation techniques for text
  readability software for adults with intellectual disabilities}. In
  \bibinfo{booktitle}{\emph{Proceedings of the 11th international ACM SIGACCESS
  conference on Computers and accessibility}}. \bibinfo{pages}{3--10}.
\newblock


\bibitem[Johann et~al\mbox{.}(2020)]%
        {Johann-et-al-2020}
\bibfield{author}{\bibinfo{person}{Verena Johann}, \bibinfo{person}{Tanja
  Könen}, {and} \bibinfo{person}{Julia Karbach}.}
  \bibinfo{year}{2020}\natexlab{}.
\newblock \showarticletitle{The unique contribution of working memory,
  inhibition, cognitive flexibility, and intelligence to reading comprehension
  and reading speed}.
\newblock \bibinfo{journal}{\emph{Child Neuropsychology}} \bibinfo{volume}{26},
  \bibinfo{number}{3} (\bibinfo{year}{2020}), \bibinfo{pages}{324--344}.
\newblock
\urldef\tempurl%
\url{https://doi.org/10.1080/09297049.2019.1649381}
\showDOI{\tempurl}
\newblock
\shownote{PMID: 31380706}.


\bibitem[Jones et~al\mbox{.}(2006)]%
        {Jones-et-al-2006}
\bibfield{author}{\bibinfo{person}{FW Jones}, \bibinfo{person}{K Long}, {and}
  \bibinfo{person}{WML Finlay}.} \bibinfo{year}{2006}\natexlab{}.
\newblock \showarticletitle{Assessing the reading comprehension of adults with
  learning disabilities}.
\newblock \bibinfo{journal}{\emph{Journal of Intellectual Disability Research}}
  \bibinfo{volume}{50}, \bibinfo{number}{6} (\bibinfo{year}{2006}),
  \bibinfo{pages}{410--418}.
\newblock


\bibitem[Kew et~al\mbox{.}(2023)]%
        {Kew-et-al-2023}
\bibfield{author}{\bibinfo{person}{Tannon Kew}, \bibinfo{person}{Alison Chi},
  \bibinfo{person}{Laura Vásquez-Rodríguez}, \bibinfo{person}{Sweta Agrawal},
  \bibinfo{person}{Dennis Aumiller}, \bibinfo{person}{Fernando Alva-Manchego},
  {and} \bibinfo{person}{Matthew Shardlow}.} \bibinfo{year}{2023}\natexlab{}.
\newblock \bibinfo{title}{BLESS: Benchmarking Large Language Models on Sentence
  Simplification}.
\newblock
\newblock
\showeprint[arxiv]{2310.15773}~[cs.CL]


\bibitem[Kim and Kim(2013)]%
        {Kim-Kim-2013}
\bibfield{author}{\bibinfo{person}{Hak~Joon Kim} {and} \bibinfo{person}{Joan
  Kim}.} \bibinfo{year}{2013}\natexlab{}.
\newblock \showarticletitle{Reading from an {LCD} monitor versus paper:
  Teenagers' reading performance}.
\newblock \bibinfo{journal}{\emph{International Journal of Research Studies in
  Educational Technology}} \bibinfo{volume}{2}, \bibinfo{number}{1}
  (\bibinfo{date}{April} \bibinfo{year}{2013}).
\newblock
\urldef\tempurl%
\url{https://doi.org/10.5861/ijrset.2012.170}
\showDOI{\tempurl}


\bibitem[Kong et~al\mbox{.}(2018)]%
        {Kong-et-al-2018}
\bibfield{author}{\bibinfo{person}{Yiren Kong}, \bibinfo{person}{Young~Sik
  Seo}, {and} \bibinfo{person}{Ling Zhai}.} \bibinfo{year}{2018}\natexlab{}.
\newblock \showarticletitle{Comparison of reading performance on screen and on
  paper: A meta-analysis}.
\newblock \bibinfo{journal}{\emph{Computers \& Education}}
  \bibinfo{volume}{123} (\bibinfo{year}{2018}), \bibinfo{pages}{138--149}.
\newblock


\bibitem[Kumar et~al\mbox{.}(2019)]%
        {Kumar-et-al-2019}
\bibfield{author}{\bibinfo{person}{Ravin Kumar}, \bibinfo{person}{Colin
  Carroll}, \bibinfo{person}{Ari Hartikainen}, {and} \bibinfo{person}{Osvaldo
  Martin}.} \bibinfo{year}{2019}\natexlab{}.
\newblock \showarticletitle{ArviZ a unified library for exploratory analysis of
  Bayesian models in Python}.
\newblock \bibinfo{journal}{\emph{Journal of Open Source Software}}
  \bibinfo{volume}{4}, \bibinfo{number}{33} (\bibinfo{year}{2019}),
  \bibinfo{pages}{1143}.
\newblock
\urldef\tempurl%
\url{https://doi.org/10.21105/joss.01143}
\showDOI{\tempurl}


\bibitem[Laban et~al\mbox{.}(2021)]%
        {Laban-et-al-2021}
\bibfield{author}{\bibinfo{person}{Philippe Laban}, \bibinfo{person}{Tobias
  Schnabel}, \bibinfo{person}{Paul Bennett}, {and} \bibinfo{person}{Marti~A.
  Hearst}.} \bibinfo{year}{2021}\natexlab{}.
\newblock \showarticletitle{Keep It Simple: Unsupervised Simplification of
  Multi-Paragraph Text}. In \bibinfo{booktitle}{\emph{Proceedings of the 59th
  Annual Meeting of the Association for Computational Linguistics and the 11th
  International Joint Conference on Natural Language Processing (Volume 1: Long
  Papers)}}. \bibinfo{publisher}{Association for Computational Linguistics},
  \bibinfo{address}{Online}, \bibinfo{pages}{6365--6378}.
\newblock
\urldef\tempurl%
\url{https://doi.org/10.18653/v1/2021.acl-long.498}
\showDOI{\tempurl}


\bibitem[Leroy et~al\mbox{.}(2013)]%
        {Leroy-et-al-2013}
\bibfield{author}{\bibinfo{person}{Gondy Leroy}, \bibinfo{person}{James~E
  Endicott}, \bibinfo{person}{David Kauchak}, \bibinfo{person}{Obay Mouradi},
  {and} \bibinfo{person}{Melissa Just}.} \bibinfo{year}{2013}\natexlab{}.
\newblock \showarticletitle{User evaluation of the effects of a text
  simplification algorithm using term familiarity on perception, understanding,
  learning, and information retention}.
\newblock \bibinfo{journal}{\emph{Journal of medical Internet research}}
  \bibinfo{volume}{15}, \bibinfo{number}{7} (\bibinfo{year}{2013}),
  \bibinfo{pages}{e144}.
\newblock


\bibitem[Linacre(1989)]%
        {Linacre-1989}
\bibfield{author}{\bibinfo{person}{John~Michael Linacre}.}
  \bibinfo{year}{1989}\natexlab{}.
\newblock \emph{\bibinfo{title}{Many-faceted Rasch measurement}}.
\newblock \bibinfo{thesistype}{Ph.\,D. Dissertation}. \bibinfo{school}{The
  University of Chicago}.
\newblock


\bibitem[Liu et~al\mbox{.}(2020)]%
        {Liu-et-al-2020}
\bibfield{author}{\bibinfo{person}{Yinhan Liu}, \bibinfo{person}{Jiatao Gu},
  \bibinfo{person}{Naman Goyal}, \bibinfo{person}{Xian Li},
  \bibinfo{person}{Sergey Edunov}, \bibinfo{person}{Marjan Ghazvininejad},
  \bibinfo{person}{Mike Lewis}, {and} \bibinfo{person}{Luke Zettlemoyer}.}
  \bibinfo{year}{2020}\natexlab{}.
\newblock \showarticletitle{Multilingual Denoising Pre-training for Neural
  Machine Translation}.
\newblock \bibinfo{journal}{\emph{Transactions of the Association for
  Computational Linguistics}}  \bibinfo{volume}{8} (\bibinfo{year}{2020}),
  \bibinfo{pages}{726--742}.
\newblock
\urldef\tempurl%
\url{https://doi.org/10.1162/tacl_a_00343}
\showDOI{\tempurl}


\bibitem[Maaß(2020)]%
        {Maass-2020}
\bibfield{author}{\bibinfo{person}{Christiane Maaß}.}
  \bibinfo{year}{2020}\natexlab{}.
\newblock \bibinfo{booktitle}{\emph{Easy Language--Plain Language--Easy
  Language Plus: Balancing comprehensibility and acceptability}}.
\newblock \bibinfo{publisher}{Frank \& Timme}.
\newblock


\bibitem[Maddela et~al\mbox{.}(2023)]%
        {Maddela-et-al-2023}
\bibfield{author}{\bibinfo{person}{Mounica Maddela}, \bibinfo{person}{Yao Dou},
  \bibinfo{person}{David Heineman}, {and} \bibinfo{person}{Wei Xu}.}
  \bibinfo{year}{2023}\natexlab{}.
\newblock \showarticletitle{{LENS}: A Learnable Evaluation Metric for Text
  Simplification}. In \bibinfo{booktitle}{\emph{Proceedings of the 61st Annual
  Meeting of the Association for Computational Linguistics (Volume 1: Long
  Papers)}}, \bibfield{editor}{\bibinfo{person}{Anna Rogers},
  \bibinfo{person}{Jordan Boyd-Graber}, {and} \bibinfo{person}{Naoaki Okazaki}}
  (Eds.). \bibinfo{publisher}{Association for Computational Linguistics},
  \bibinfo{address}{Toronto, Canada}, \bibinfo{pages}{16383--16408}.
\newblock
\urldef\tempurl%
\url{https://doi.org/10.18653/v1/2023.acl-long.905}
\showDOI{\tempurl}


\bibitem[Mallinson et~al\mbox{.}(2020)]%
        {Mallinson-et-al-2020}
\bibfield{author}{\bibinfo{person}{Jonathan Mallinson}, \bibinfo{person}{Rico
  Sennrich}, {and} \bibinfo{person}{Mirella Lapata}.}
  \bibinfo{year}{2020}\natexlab{}.
\newblock \showarticletitle{Zero-Shot Crosslingual Sentence Simplification}. In
  \bibinfo{booktitle}{\emph{Proceedings of the 2020 Conference on Empirical
  Methods in Natural Language Processing (EMNLP)}}.
  \bibinfo{publisher}{Association for Computational Linguistics},
  \bibinfo{address}{Online}, \bibinfo{pages}{5109--5126}.
\newblock
\urldef\tempurl%
\url{https://doi.org/10.18653/v1/2020.emnlp-main.415}
\showDOI{\tempurl}


\bibitem[Martin et~al\mbox{.}(2022)]%
        {Martin-et-al-2022}
\bibfield{author}{\bibinfo{person}{Louis Martin}, \bibinfo{person}{Angela Fan},
  \bibinfo{person}{{\'E}ric de~la Clergerie}, \bibinfo{person}{Antoine Bordes},
  {and} \bibinfo{person}{Beno{\^\i}t Sagot}.} \bibinfo{year}{2022}\natexlab{}.
\newblock \showarticletitle{{MUSS}: Multilingual Unsupervised Sentence
  Simplification by Mining Paraphrases}. In
  \bibinfo{booktitle}{\emph{Proceedings of the Thirteenth Language Resources
  and Evaluation Conference}}. \bibinfo{publisher}{European Language Resources
  Association}, \bibinfo{address}{Marseille, France},
  \bibinfo{pages}{1651--1664}.
\newblock
\urldef\tempurl%
\url{https://aclanthology.org/2022.lrec-1.176}
\showURL{%
\tempurl}


\bibitem[{Menéndez Álvarez-Dardet} et~al\mbox{.}(2020)]%
        {Mendenez-et-al-2020}
\bibfield{author}{\bibinfo{person}{Susana {Menéndez Álvarez-Dardet}},
  \bibinfo{person}{Bárbara {Lorence Lara}}, {and} \bibinfo{person}{Javier
  Pérez-Padilla}.} \bibinfo{year}{2020}\natexlab{}.
\newblock \showarticletitle{Older adults and ICT adoption: Analysis of the use
  and attitudes toward computers in elderly Spanish people}.
\newblock \bibinfo{journal}{\emph{Computers in Human Behavior}}
  \bibinfo{volume}{110} (\bibinfo{year}{2020}), \bibinfo{pages}{106377}.
\newblock
\showISSN{0747-5632}
\urldef\tempurl%
\url{https://doi.org/10.1016/j.chb.2020.106377}
\showDOI{\tempurl}


\bibitem[Meyer and Schvaneveldt(1971)]%
        {Meyer-Schvaneveldt-1971}
\bibfield{author}{\bibinfo{person}{David~E. Meyer} {and}
  \bibinfo{person}{Roger~W. Schvaneveldt}.} \bibinfo{year}{1971}\natexlab{}.
\newblock \showarticletitle{Facilitation in recognizing pairs of words:
  Evidence of a dependence between retrieval operations.}
\newblock \bibinfo{journal}{\emph{Journal of Experimental Psychology}}
  \bibinfo{volume}{90}, \bibinfo{number}{2} (\bibinfo{date}{Oct.}
  \bibinfo{year}{1971}), \bibinfo{pages}{227--234}.
\newblock
\urldef\tempurl%
\url{https://doi.org/10.1037/h0031564}
\showDOI{\tempurl}


\bibitem[Micai et~al\mbox{.}(2019)]%
        {Micai-et-al-2019}
\bibfield{author}{\bibinfo{person}{Martina Micai}, \bibinfo{person}{Mila
  Vulchanova}, {and} \bibinfo{person}{David Salda{\~{n}}a}.}
  \bibinfo{year}{2019}\natexlab{}.
\newblock \showarticletitle{Do Individuals with Autism Change Their Reading
  Behavior to Adapt to Errors in the Text?}
\newblock \bibinfo{journal}{\emph{Journal of Autism and Developmental
  Disorders}} \bibinfo{volume}{49}, \bibinfo{number}{10} (\bibinfo{date}{July}
  \bibinfo{year}{2019}), \bibinfo{pages}{4232--4243}.
\newblock
\urldef\tempurl%
\url{https://doi.org/10.1007/s10803-019-04108-8}
\showDOI{\tempurl}


\bibitem[Pappert and Bock(2019)]%
        {Pappert-Bock-2019}
\bibfield{author}{\bibinfo{person}{Sandra Pappert} {and}
  \bibinfo{person}{Bettina~M Bock}.} \bibinfo{year}{2019}\natexlab{}.
\newblock \showarticletitle{Easy-to-read {G}erman put to the test: Do adults
  with intellectual disability or functional illiteracy benefit from compound
  segmentation?}
\newblock \bibinfo{journal}{\emph{Reading and Writing}} (\bibinfo{year}{2019}),
  \bibinfo{pages}{1--27}.
\newblock


\bibitem[Ramsten et~al\mbox{.}(2018)]%
        {Ramsten-et-al-2018}
\bibfield{author}{\bibinfo{person}{Camilla Ramsten},
  \bibinfo{person}{Lene~Karine Martin}, \bibinfo{person}{Munir Dag}, {and}
  \bibinfo{person}{Lena~Marmst{\aa}l Hammar}.} \bibinfo{year}{2018}\natexlab{}.
\newblock \showarticletitle{Information and communication technology use in
  daily life among young adults with mild-to-moderate intellectual disability}.
\newblock \bibinfo{journal}{\emph{Journal of Intellectual Disabilities}}
  \bibinfo{volume}{24} (\bibinfo{year}{2018}), \bibinfo{pages}{289--308}.
\newblock


\bibitem[Reitan and Wolfson(1993)]%
        {Reitan-Wolfson-1993}
\bibfield{author}{\bibinfo{person}{Ralph Reitan} {and} \bibinfo{person}{Deborah
  Wolfson}.} \bibinfo{year}{1993}\natexlab{}.
\newblock \bibinfo{booktitle}{\emph{The {H}alstead-{R}eitan neuropsychological
  test battery: Theory and clinical interpretation}}.
\newblock \bibinfo{publisher}{Neuropsychology Press}.
\newblock


\bibitem[Rello et~al\mbox{.}(2013)]%
        {Rello-et-al-2013b}
\bibfield{author}{\bibinfo{person}{Luz Rello}, \bibinfo{person}{Ricardo
  Baeza-Yates}, \bibinfo{person}{Stefan Bott}, {and} \bibinfo{person}{Horacio
  Saggion}.} \bibinfo{year}{2013}\natexlab{}.
\newblock \showarticletitle{Simplify or help? Text simplification strategies
  for people with dyslexia}. In \bibinfo{booktitle}{\emph{Proceedings of the
  10th International Cross-Disciplinary Conference on Web Accessibility}}.
  \bibinfo{pages}{1--10}.
\newblock


\bibitem[Riddell et~al\mbox{.}(2021)]%
        {pystan}
\bibfield{author}{\bibinfo{person}{Allen Riddell}, \bibinfo{person}{Ari
  Hartikainen}, {and} \bibinfo{person}{Matthew Carter}.}
  \bibinfo{year}{2021}\natexlab{}.
\newblock \bibinfo{title}{pystan}.
\newblock \bibinfo{howpublished}{PyPI}.
\newblock


\bibitem[Rios et~al\mbox{.}(2021)]%
        {Rios-et-al-2021}
\bibfield{author}{\bibinfo{person}{Annette Rios}, \bibinfo{person}{Nicolas
  Spring}, \bibinfo{person}{Tannon Kew}, \bibinfo{person}{Marek Kostrzewa},
  \bibinfo{person}{Andreas S{\"a}uberli}, \bibinfo{person}{Mathias M{\"u}ller},
  {and} \bibinfo{person}{Sarah Ebling}.} \bibinfo{year}{2021}\natexlab{}.
\newblock \showarticletitle{A New Dataset and Efficient Baselines for
  Document-level Text Simplification in {G}erman}. In
  \bibinfo{booktitle}{\emph{Proceedings of the Third Workshop on New Frontiers
  in Summarization}}. \bibinfo{publisher}{Association for Computational
  Linguistics}, \bibinfo{address}{Online and in Dominican Republic},
  \bibinfo{pages}{152--161}.
\newblock
\urldef\tempurl%
\url{https://doi.org/10.18653/v1/2021.newsum-1.16}
\showDOI{\tempurl}


\bibitem[Ryan et~al\mbox{.}(2023)]%
        {Ryan-et-al-2023}
\bibfield{author}{\bibinfo{person}{Michael Ryan}, \bibinfo{person}{Tarek
  Naous}, {and} \bibinfo{person}{Wei Xu}.} \bibinfo{year}{2023}\natexlab{}.
\newblock \showarticletitle{Revisiting non-{E}nglish Text Simplification: A
  Unified Multilingual Benchmark}. In \bibinfo{booktitle}{\emph{Proceedings of
  the 61st Annual Meeting of the Association for Computational Linguistics
  (Volume 1: Long Papers)}}. \bibinfo{publisher}{Association for Computational
  Linguistics}, \bibinfo{address}{Toronto, Canada},
  \bibinfo{pages}{4898--4927}.
\newblock
\urldef\tempurl%
\url{https://doi.org/10.18653/v1/2023.acl-long.269}
\showDOI{\tempurl}


\bibitem[Saggion et~al\mbox{.}(2015)]%
        {Saggion-et-al-2015}
\bibfield{author}{\bibinfo{person}{Horacio Saggion}, \bibinfo{person}{Sanja
  \v{S}tajner}, \bibinfo{person}{Stefan Bott}, \bibinfo{person}{Simon Mille},
  \bibinfo{person}{Luz Rello}, {and} \bibinfo{person}{Biljana Drndarevic}.}
  \bibinfo{year}{2015}\natexlab{}.
\newblock \showarticletitle{Making It Simplext: Implementation and Evaluation
  of a Text Simplification System for Spanish}.
\newblock \bibinfo{journal}{\emph{ACM Trans. Access. Comput.}}
  \bibinfo{volume}{6}, \bibinfo{number}{4}, Article \bibinfo{articleno}{14}
  (\bibinfo{date}{may} \bibinfo{year}{2015}), \bibinfo{numpages}{36}~pages.
\newblock
\showISSN{1936-7228}
\urldef\tempurl%
\url{https://doi.org/10.1145/2738046}
\showDOI{\tempurl}


\bibitem[Saletta and Winberg(2019)]%
        {Saletta-Winberg-2019}
\bibfield{author}{\bibinfo{person}{Meredith Saletta} {and}
  \bibinfo{person}{Jennifer Winberg}.} \bibinfo{year}{2019}\natexlab{}.
\newblock \showarticletitle{Leveled Texts for Adults With Intellectual or
  Developmental Disabilities: A Pilot Study}.
\newblock \bibinfo{journal}{\emph{Focus on Autism and Other Developmental
  Disabilities}} \bibinfo{volume}{34}, \bibinfo{number}{2}
  (\bibinfo{year}{2019}), \bibinfo{pages}{118--127}.
\newblock
\urldef\tempurl%
\url{https://doi.org/10.1177/1088357618803332}
\showDOI{\tempurl}
\showeprint{https://doi.org/10.1177/1088357618803332}


\bibitem[Samejima(1997)]%
        {Samejima-1997}
\bibfield{author}{\bibinfo{person}{Fumiko Samejima}.}
  \bibinfo{year}{1997}\natexlab{}.
\newblock \bibinfo{booktitle}{\emph{Graded Response Model}}.
\newblock \bibinfo{publisher}{Springer New York}, \bibinfo{address}{New York,
  NY}, \bibinfo{pages}{85--100}.
\newblock
\showISBNx{978-1-4757-2691-6}
\urldef\tempurl%
\url{https://doi.org/10.1007/978-1-4757-2691-6_5}
\showDOI{\tempurl}


\bibitem[Schiffl(2020)]%
        {Schiffl-2020}
\bibfield{author}{\bibinfo{person}{Laura Schiffl}.}
  \bibinfo{year}{2020}\natexlab{}.
\newblock \showarticletitle{Hierarchies in Lexical Complexity: Do Effects of
  Word Frequency, Word Length and Repetition Exist for the Visual Word
  Processing of People with Cognitive Impairments?}
\newblock In \bibinfo{booktitle}{\emph{Easy Language Research: Text and User
  Perspectives}}, \bibfield{editor}{\bibinfo{person}{Silvia Hansen-Schirra}
  {and} \bibinfo{person}{Christiane Maaß}} (Eds.). \bibinfo{publisher}{Frank
  \& Timme GmbH}, \bibinfo{pages}{227--239}.
\newblock


\bibitem[{\v{S}}tajner and Nisioi(2018)]%
        {Stajner-Nisioi-2018}
\bibfield{author}{\bibinfo{person}{Sanja {\v{S}}tajner} {and}
  \bibinfo{person}{Sergiu Nisioi}.} \bibinfo{year}{2018}\natexlab{}.
\newblock \showarticletitle{A Detailed Evaluation of Neural
  Sequence-to-Sequence Models for In-domain and Cross-domain Text
  Simplification}. In \bibinfo{booktitle}{\emph{Proceedings of the Eleventh
  International Conference on Language Resources and Evaluation ({LREC}
  2018)}}. \bibinfo{publisher}{European Language Resources Association (ELRA)},
  \bibinfo{address}{Miyazaki, Japan}.
\newblock
\urldef\tempurl%
\url{https://www.aclweb.org/anthology/L18-1479}
\showURL{%
\tempurl}


\bibitem[Stodden(2021)]%
        {Stodden-2021}
\bibfield{author}{\bibinfo{person}{Regina Stodden}.}
  \bibinfo{year}{2021}\natexlab{}.
\newblock \showarticletitle{When the Scale is Unclear - Analysis of the
  Interpretation of Rating Scales in Human Evaluation of Text Simplification}.
  In \bibinfo{booktitle}{\emph{Proceedings of the First Workshop on Current
  Trends in Text Simplification {(CTTS} 2021) co-located with the 37th
  Conference of the Spanish Society for Natural Language Processing
  (SEPLN2021), Online (initially located in M{\'{a}}laga, Spain), September
  21st, 2021}} \emph{(\bibinfo{series}{{CEUR} Workshop Proceedings},
  Vol.~\bibinfo{volume}{2944})}, \bibfield{editor}{\bibinfo{person}{Horacio
  Saggion}, \bibinfo{person}{Sanja {\v{S}}tajner}, \bibinfo{person}{Daniel
  Ferr{\'{e}}s}, {and} \bibinfo{person}{Kim~Cheng Sheang}} (Eds.).
  \bibinfo{publisher}{CEUR-WS.org}.
\newblock
\urldef\tempurl%
\url{http://ceur-ws.org/Vol-2944/paper6.pdf}
\showURL{%
\tempurl}


\bibitem[Støle et~al\mbox{.}(2020)]%
        {Stole-et-al-2020}
\bibfield{author}{\bibinfo{person}{Hildegunn Støle}, \bibinfo{person}{Anne
  Mangen}, {and} \bibinfo{person}{Knut Schwippert}.}
  \bibinfo{year}{2020}\natexlab{}.
\newblock \showarticletitle{Assessing children's reading comprehension on paper
  and screen: A mode-effect study}.
\newblock \bibinfo{journal}{\emph{Computers \& Education}}
  \bibinfo{volume}{151} (\bibinfo{year}{2020}), \bibinfo{pages}{103861}.
\newblock
\showISSN{0360-1315}
\urldef\tempurl%
\url{https://doi.org/10.1016/j.compedu.2020.103861}
\showDOI{\tempurl}


\bibitem[Säuberli et~al\mbox{.}(2023)]%
        {Saeuberli-et-al-2023}
\bibfield{author}{\bibinfo{person}{Andreas Säuberli}, \bibinfo{person}{Silvia
  Hansen-Schirra}, \bibinfo{person}{Franz Holzknecht}, \bibinfo{person}{Silke
  Gutermuth}, \bibinfo{person}{Silvana Deilen}, \bibinfo{person}{Laura
  Schiffl}, {and} \bibinfo{person}{Sarah Ebling}.}
  \bibinfo{year}{2023}\natexlab{}.
\newblock \showarticletitle{Enabling text comprehensibility assessment for
  people with intellectual disabilities using a mobile application}.
\newblock \bibinfo{journal}{\emph{Frontiers in Communication}}
  \bibinfo{volume}{8} (\bibinfo{year}{2023}).
\newblock
\showISSN{2297-900X}
\urldef\tempurl%
\url{https://doi.org/10.3389/fcomm.2023.1175625}
\showDOI{\tempurl}


\bibitem[Trauzettel-Klosinski et~al\mbox{.}(2012)]%
        {Trauzettel-Klosinski-et-al-2012}
\bibfield{author}{\bibinfo{person}{Susanne Trauzettel-Klosinski},
  \bibinfo{person}{Klaus Dietz}, {and} \bibinfo{person}{the IReST
  Study~Group}.} \bibinfo{year}{2012}\natexlab{}.
\newblock \showarticletitle{{Standardized Assessment of Reading Performance:
  The New International Reading Speed Texts IReST}}.
\newblock \bibinfo{journal}{\emph{Investigative Ophthalmology \& Visual
  Science}} \bibinfo{volume}{53}, \bibinfo{number}{9} (\bibinfo{date}{08}
  \bibinfo{year}{2012}), \bibinfo{pages}{5452--5461}.
\newblock
\showISSN{1552-5783}
\urldef\tempurl%
\url{https://doi.org/10.1167/iovs.11-8284}
\showDOI{\tempurl}


\bibitem[Van~der Linden et~al\mbox{.}(1999)]%
        {VanDerLinden-et-al-1999}
\bibfield{author}{\bibinfo{person}{Wim~J Van~der Linden},
  \bibinfo{person}{David~J Scrams}, {and} \bibinfo{person}{Deborah~L
  Schnipke}.} \bibinfo{year}{1999}\natexlab{}.
\newblock \showarticletitle{Using response-time constraints to control for
  differential speededness in computerized adaptive testing}.
\newblock \bibinfo{journal}{\emph{Applied psychological measurement}}
  \bibinfo{volume}{23}, \bibinfo{number}{3} (\bibinfo{year}{1999}),
  \bibinfo{pages}{195--210}.
\newblock


\bibitem[\v{S}tajner(2021)]%
        {Stajner-2021}
\bibfield{author}{\bibinfo{person}{Sanja \v{S}tajner}.}
  \bibinfo{year}{2021}\natexlab{}.
\newblock \showarticletitle{Automatic text simplification for social good:
  Progress and challenges}. In \bibinfo{booktitle}{\emph{Findings of the
  Association for Computational Linguistics: ACL-IJCNLP 2021}}.
  \bibinfo{publisher}{Association for Computational Linguistics},
  \bibinfo{address}{Online}, \bibinfo{pages}{2637--2652}.
\newblock
\urldef\tempurl%
\url{https://doi.org/10.18653/v1/2021.findings-acl.233}
\showDOI{\tempurl}


\bibitem[Wallot et~al\mbox{.}(2014)]%
        {Wallot-et-al-2014}
\bibfield{author}{\bibinfo{person}{Sebastian Wallot}, \bibinfo{person}{Beth~A.
  O’Brien}, \bibinfo{person}{Anna Haussmann}, \bibinfo{person}{Heidi Kloos},
  {and} \bibinfo{person}{Marlene~S. Lyby}.} \bibinfo{year}{2014}\natexlab{}.
\newblock \showarticletitle{The role of reading time complexity and reading
  speed in text comprehension.}
\newblock \bibinfo{journal}{\emph{Journal of Experimental Psychology: Learning,
  Memory, and Cognition}} \bibinfo{volume}{40}, \bibinfo{number}{6}
  (\bibinfo{year}{2014}), \bibinfo{pages}{1745--1765}.
\newblock
\showISSN{0278-7393}
\urldef\tempurl%
\url{https://doi.org/10.1037/xlm0000030}
\showDOI{\tempurl}


\bibitem[Wechsler(2003)]%
        {Wechsler-2003}
\bibfield{author}{\bibinfo{person}{David Wechsler}.}
  \bibinfo{year}{2003}\natexlab{}.
\newblock \bibinfo{title}{Wechsler Intelligence Scale for Children, Fourth
  Edition}.
\newblock
\newblock
\urldef\tempurl%
\url{https://doi.org/10.1037/t15174-000}
\showDOI{\tempurl}


\bibitem[You(2022)]%
        {You-2022}
\bibfield{author}{\bibinfo{person}{Hyesun You}.}
  \bibinfo{year}{2022}\natexlab{}.
\newblock \showarticletitle{Bayesian Versus Frequentist Estimation for Item
  Response Theory Models of Interdisciplinary Science Assessment}.
\newblock \bibinfo{journal}{\emph{Interdisciplinary Journal of Environmental
  and Science Education}} \bibinfo{volume}{18}, \bibinfo{number}{4}
  (\bibinfo{date}{jul} \bibinfo{year}{2022}), \bibinfo{pages}{e2297}.
\newblock
\urldef\tempurl%
\url{https://doi.org/10.21601/ijese/12299}
\showDOI{\tempurl}


\end{thebibliography}

\appendix

\section{Priors for model parameters}

\begin{table}[h!]
  \centering
  \begin{tabular}{lll}
    \toprule
    Measurement & Parameter & Prior distribution \\
    \midrule
    Comprehension question
      & $\mu \in \mathbb{R}$ & $\mathcal{N}(0, 1)$ \\
    accuracy
      & $\alpha \in \mathbb{R}^{16}$ & $\mathcal{N}(0, 1)$ \\
      & $\beta \in \mathbb{R}^{48}$ & $\mathcal{N}(0, 1)$ \\
      & $\delta \in \mathbb{R}^3$ & $\mathcal{N}(0, 1)$ \\
    \midrule
    Perceived difficulty
      & $\mu \in \mathbb{R}$ & $\mathcal{N}(0, 1)$ \\
    ratings
      & $\alpha \in \mathbb{R}^{16}$ & $\mathcal{N}(0, 1)$ \\
      & $\gamma \in \mathbb{R}^{12}$ & $\mathcal{N}(0, 1)$ \\
      & $\delta \in \mathbb{R}^3$ & $\mathcal{N}(0, 1)$ \\
      & $c_k - c_{k-1} \in \mathbb{R}^3$ & $\mathcal{N}(0, 1)$ \\
    \midrule
    Response time
      & $\mu \in \mathbb{R}$ & $\mathcal{N}(0, 5)$ \\
      & $\sigma \in \mathbb{R}$ & $\Gamma(1, 5)$ \\
      & $\alpha \in \mathbb{R}^{16}$ & $\mathcal{N}(0, 5)$ \\
      & $\beta \in \mathbb{R}^{48}$ & $\mathcal{N}(0, 5)$ \\
      & $\delta \in \mathbb{R}^3$ & $\mathcal{N}(0, 5)$ \\
    \midrule
    Reading speed
      & $\mu \in \mathbb{R}$ & $\mathcal{N}(0, 5)$ \\
      & $\sigma \in \mathbb{R}$ & $\Gamma(1, 5)$ \\
      & $\alpha \in \mathbb{R}^{16}$ & $\mathcal{N}(0, 5)$ \\
      & $\gamma \in \mathbb{R}^{12}$ & $\mathcal{N}(0, 5)$ \\
      & $\delta \in \mathbb{R}^3$ & $\mathcal{N}(0, 5)$ \\
    \bottomrule
  \end{tabular}
  \caption{Overview of the prior distributions used for the Bayesian models}
  \label{tab:priors}
\end{table}

Table \ref{tab:priors} shows the prior distributions we chose for all model parameters.

Note about $c_k$: For five rating categories, there are four threshold parameters in the graded response model. They need to be in ascending order and sum to zero. Therefore, instead of sampling the threshold parameters directly, we sample the differences between adjacent threshold parameters, of which there are three. For all the other parameters, the sum-to-zero constraint is achieved by dividing by the mean. For implementation details, refer to the \emph{Stan} code in the supplementary material.
\end{document}